\documentclass[times,twocolumn,final]{elsarticle}

\usepackage{medima}
\usepackage{framed,multirow}

\usepackage{amssymb}
\usepackage{latexsym}

\usepackage{url}
\usepackage{xcolor}
\usepackage{array}

\usepackage{hyperref}
\hypersetup{colorlinks=true}

\usepackage{graphicx}
\usepackage{amsmath,amssymb} 
\usepackage{color}
\usepackage[font=small,labelfont=bf]{caption}

\usepackage{makeidx}  
\usepackage{color}
\usepackage{graphicx}

\usepackage{xcolor}
\usepackage{colortbl}
\definecolor{maroon}{cmyk}{0,0.87,0.68,0.32}

\usepackage{amsmath}
\usepackage{amssymb}
\usepackage[labelformat=simple]{subfig}
\usepackage[ruled,lined,linesnumbered]{algorithm2e}
\usepackage{algorithmic}
\usepackage{blindtext}
\usepackage{enumitem}
\usepackage{multirow}
\usepackage{threeparttable/threeparttable}
\usepackage{bm}
\usepackage[normalem]{ulem}
\DeclareMathOperator*{\argmax}{argmax}
\usepackage{ulem}

\usepackage{amssymb}
\usepackage{pifont}

\newcommand{\ie}{\mbox{\emph{i.e.,\ }}}
\newcommand{\eg}{\mbox{\emph{e.g.,\ }}}

\newcommand{\tabincell}[2]{\begin{tabular}{@{}#1@{}}#2\end{tabular}}

\newcolumntype{P}[1]{>{\centering\arraybackslash}p{#1}}

\definecolor{maroon}{cmyk}{0,0.87,0.68,0.32}
\definecolor{fgreen}{rgb}{0.,0.58,0.2}
\definecolor{iblue}{rgb}{0.06, 0.75, 1.0}
\definecolor{revblue}{rgb}{0.0, 0.44, 1.0}
\definecolor{brightpink}{rgb}{1.0, 0.2, 0.4}
\newenvironment{jlblue}{\color{black}}{}

\usepackage{soul}

\definecolor{newcolor}{rgb}{.8,.349,.1}

\journal{Medical Image Analysis}

\begin{document}

\verso{Zongwei Zhou \textit{et~al.}}

\begin{frontmatter}

\title{Active, Continual Fine Tuning of Convolutional Neural Networks\\ for Reducing Annotation Efforts}

\author[1]{Zongwei \snm{Zhou}}
\author[1]{Jae Y. \snm{Shin}}
\author[2]{Suryakanth R. \snm{Gurudu}}
\author[3]{Michael B. \snm{Gotway}}
\author[1]{Jianming \snm{Liang}\corref{cor1}}
\cortext[cor1]{Corresponding author: \href{mailto:Jianming.Liang@asu.edu}{Jianming.Liang@asu.edu} (Jianming Liang)}

\address[1]{Department of medical Informatics, Arizona State University, Scottsdale, AZ 85259, USA}
\address[2]{Division of Gastroenterology and Hepatology, Mayo Clinic, Scottsdale, AZ 85259, USA}
\address[3]{Department of Radiology, Mayo Clinic, Scottsdale, AZ 85259, USA}

\received{***}
\finalform{***}
\accepted{***}
\availableonline{***}
\communicated{***}

\begin{abstract}
The splendid success of convolutional neural networks (CNNs) in computer vision is largely attributable to the availability of massive annotated datasets, such as \textsc{ImageNet} and \textsc{Places}. However, in medical imaging, it is challenging to create such large annotated datasets, as annotating medical images is not only tedious, laborious, and time consuming, but it also demands costly, specialty-oriented skills, which are not easily accessible. To dramatically reduce annotation cost, this paper presents a novel method to naturally integrate active learning and transfer learning (fine-tuning) into a single framework, which starts directly with a pre-trained CNN to seek ``worthy''  samples for annotation and gradually enhances the (fine-tuned) CNN via continual fine-tuning. We have evaluated our method using three distinct medical imaging applications, demonstrating that it can reduce annotation efforts by at least half compared with random selection.
\end{abstract}


\end{frontmatter}


\section{Introduction}
\label{sec:introduction}

Convolutional neural networks (CNNs)~\citep{lecun2015deep} have ushered in a revolution in computer vision owing to the use of large annotated datasets, such as \textsc{ImageNet}~\citep{deng2009imagenet} and \textsc{Places}~\citep{zhou2017places}. As evidenced by two recent books~\citep{shen2019medical,zhou2019handbook} and numerous compelling techniques for different imaging tasks~\citep{moen2019deep,yamamoto2019automated,ravizza2019predicting,esteva2019guide,huang2020penet,isensee2021nnu}, there is widespread and intense interest in applying CNNs to medical image analysis, but the adoption of CNNs in medical imaging is hampered by the lack of such large annotated datasets. Annotating medical images is not only tedious and time consuming, but it also requires costly, specialty-oriented knowledge and skills, which are not readily accessible. Therefore, we seek to answer this critical question: {\em How to dramatically reduce the cost of annotation when applying CNNs to medical imaging}?
In doing so, we have developed a novel method called ACFT (active, continual fine-tuning) to naturally integrate active learning and transfer learning into a single framework. Our ACFT method starts directly with a pre-trained CNN to seek ``salient'' samples from the unannotated pool for annotation, and the (fine-tuned) CNN is continually fine-tuned using newly annotated samples combined with all misclassified samples. We have evaluated our method in three different applications, including colonoscopy frame classification, polyp detection, and pulmonary embolism (PE) detection, demonstrating that the cost of annotation can be reduced by at least half.

This performance is attributable to a simple yet powerful observation: to boost the performance of CNNs in medical imaging, multiple patches are usually generated automatically for each sample through data augmentation; these patches generated from the same sample share the same label, and are naturally expected to have similar predictions by the current CNN before they are expanded into the training dataset. As a result, their {\em entropy}~\citep{shannon1948mathematical} and {\em diversity}~\citep{kukar2003transductive} provide a useful indicator of the ``power'' of a sample for elevating the performance of the current CNN. However, automatic data augmentation inevitably generates ``hard'' samples, injecting noisy labels. Therefore, to significantly enhance the robustness of active selection, we compute entropy and diversity from only a portion of the patches according to the majority predictions detailed in Sec.~\ref{sec:majority_selection}) by the current CNN. 
Furthermore, to strike a balance between exploration and exploitation, we incorporate randomness in our active selection as detailed in Sec.~\ref{sec:randomness}; and to prevent catastrophic forgetting, we combine newly selected samples with misclassified samples as described in Sec.~\ref{sec:comparison_learning_strategies}.
    
Several researchers have demonstrated the utility of fine-tuning CNNs for medical image analysis, but they only performed one-time fine-tuning; that is, simply fine-tuning a pre-trained CNN once with all available training samples, involving no active selection processes \citep{tajbakhsh2016convolutional,lu2017deep,esteva2017dermatologist,mormont2018comparison,ding2018deep,irvin2019chexpert,zhou2019models,chen2019med3d,tajbakhsh2019computer,ardila2019end}. To our knowledge, our proposed method is among the first to integrate active learning into fine-tuning CNNs in a continual fashion to make CNNs more amenable to medical image analysis, particularly with the intention of decreasing the efforts of annotation dramatically.  Compared with conventional active learning, our method, summarized as Alg.~\ref{alg:ACFT},  offers eight {\bf advantages}:
\begin{enumerate}[noitemsep]
    \item Our algorithm starts with a completely empty labeled dataset, requiring no seed-labeled samples (see Alg.~\ref{alg:ACFT});
    \item Our algorithm actively selects the most informative and representative samples by naturally exploiting expected consistency among the patches within each sample (see Sec.~\ref{sec:selection_illustrated});
    \item Our algorithm computes selection criteria locally on a small number of patches within each sample, saving considerable computation time (see Sec.~\ref{sec:entropy_diversity});
    \item Our algorithm automatically handles noisy labels via majority selection (see Sec.~\ref{sec:majority_selection});
    \item Our algorithm balances exploration and exploitation by incorporating randomness into active selection (see Sec.~\ref{sec:randomness}).
    \item Our algorithm incrementally improves the learner through continual fine-tuning rather than through repeated re-training (see Sec.~\ref{sec:observation_criteria});
    \item Our algorithm focuses on hard samples, preventing catastrophic forgetting (see Sec.~\ref{sec:comparison_learning_strategies});
    \item Our algorithm autonomously balances training samples among classes (see Sec.~\ref{sec:automatically_balancing} and \figurename~\ref{fig:balance_ratio});
\end{enumerate}
More importantly, our method has the potential to positively impact computer-aided diagnosis (CAD) in medical imaging. The current regulations require that CAD systems be deployed in a ``closed'' environment, in which all CAD results are reviewed and errors, if any, must be corrected by radiologists. As a result, all false positives are dismissed and all false negatives are supplied, an instant on-line feedback process that makes it possible for CAD systems to be self-learning and self-improving after deployment given the continual fine-tuning capability of our method.

\section{Related work}
\label{sec:related_work}

\subsection{Our work}
\label{sec:our_work}

We presented AIFT (active, incremental fine-tuning) in our CVPR paper~\citep{zhou2017fine} to integrate active learning and deep learning via continual fine-tuning. Nevertheless, AIFT was limited to binary classifications and medical imaging, and used all labeled samples available at each step, thereby demanding extensive training time and substantial computer memory. Our current approach is a significant extension of our previous work with several major enhancements: (1) generalization from binary classification to multi-class classification; (2) extension from computer-aided diagnosis in medical imaging to scene classification in natural images; (3) combination of newly selected samples with hard (misclassified) ones, to eliminate easy samples for reducing training time, and to concentrate on hard samples for preventing catastrophic forgetting; (4) injection of randomness to enhance robustness in active selection; (5) extensive experimentation with all reasonable combinations of data and models in search of an optimal strategy; (6) demonstration of consistent annotation reduction using different CNN architectures; and (7) illustration of the active selection process using a gallery of patches associated with predictions.

\subsection{Transfer learning for medical imaging}
\label{sec:transfer_learning}

Pre-training a model on large-scale image datasets and then fine-tuning it on various target tasks has become a \textit{de facto} paradigm across many medical specialties. As summarized by~\citet{irvin2019chexpert}, to classify the common thoracic diseases on chest radiography, nearly all the leading approaches~\citep{guan2018multi,guendel2018learning,tang2018attention,ma2019multi} follow this paradigm by adopting different architectures along with their weights pre-trained from ImageNet. Other representative medical applications include identifying skin cancer from dermatologist level photographs~\citep{esteva2017dermatologist}, diagnosing Alzheimer's Disease~\citep{ding2018deep} from $^{18}$F-FDG PET of the brain, and performing effective detection of pulmonary embolism~\citep{tajbakhsh2019computer} from CTPA. 
Recent breakthrough in self-supervised pre-training~\citep{grill2020bootstrap,caron2020unsupervised,chen2020exploring}, on the other hand, has led to visual representation that approaches and possibly surpasses what was learned from ImageNet. Self-supervised pre-training has also been adopted for the medical domain, wherein \citet{zhou2019models,zhu2020rubik,feng2020parts2whole,haghighi2020learning,azizi2021big} develop generic CNNs that are directly pre-trained from medical images, mitigating the mandatory requirement of expert annotation and reducing the large domain gap between natural and medical images. Despite the immense popularity of transfer learning in medical imaging, these works exclusively employed {\em one-time fine-tuning}---simply fine-tuning a pre-trained CNN with available training samples for only one time. In real-world applications, instead of training on a still dataset, experts record new samples constantly and expect the samples to be used upon their availability; with the ability to deal with new data, continual learning is the bridge to active and open world learning~\citep{mundt2020wholistic}. Compared with the existing continual learning approaches~\citep{kading2016fine,zhou2017fine}, our newly devised learning strategy is more amenable to active fine-tuning because it focuses more on the newly annotated samples and also recognizes those misclassified ones, eliminating repeated training on those easy samples in the annotated pool.

\subsection{Integrating active learning with deep learning}
\label{sec:active_learning}

The uncertainty and diversity are the most compelling active selection criteria, which appraise the worthiness of annotating a sample from two different aspects. Uncertainty-based criteria argue that the more uncertain a prediction is, the more value added when including the label of that sample into the training set. Sampling with least confidence~\citep{culotta2005reducing}, large entropy~\citep{dagan1995committee,mahapatra2018efficient,shao2018deep,kuo2018cost}, or margins~\citep{scheffer2001active,balcan2007margin} of the prediction has been successful in training models with fewer labels than random sampling. The limitation of uncertainty-based criteria is that some of the selected samples are prone to redundancy and outliers~\citep{sourati2019intelligent} and may not be representative enough for the data distribution as a whole. Alternatively, diversity-based criteria have the advantage of selecting a set of most representative samples, related to the labeled ones, from those in the rest of the unlabeled set. The intuition is that there is no need to repeatedly annotate those samples with context information if the most representative one has already been covered. Mutual information~\citep{li2013adaptive,gal2017deep}, Kullback-Leibler divergence~\citep{kulick2014active,mccallumzy1998employing}, Fisher information~\citep{sourati2018active,sourati2019intelligent}, K-centers and core sets~\citep{sener2017active}, calculated among either model predictions or image features, are often used to ensure the diversity. Although alleviating redundancy and outliers, a serious hurdle of diversity-based criteria is the computational complexity for a large pool of unlabeled samples. We address this issue by measuring diversity over patches augmented from the same sample, making the calculation much more manageable. To exploit the benefits and potentials of the two selecting aspects, the studies of \citet{wang2018deep,ozdemir2018active,mahapatra2018efficient,shui2020deep}, as well as our ACFT, consider the mixture strategy of combing uncertainty and diversity explicitly. \citet{yang2017suggestive,beluch2018power,kuo2018cost} further compute the selection criteria from an ensemble of CNNs---these approaches are, however, very costly in computation, as they must train a set of models to compute their uncertainty measure based on models' disagreements. For additional active learning methods, we refer the reader to comprehensive literature reviews~\citep{tajbakhsh2020embracing,munjal2020towards,hino2020active,ren2020survey}; but these existing methods are fundamentally different from our ACFT in that they all repeatedly re-trained CNNs from scratch at each step, whereas we continually fine-tune the (fine-tuned) CNN incrementally. As a result, our ACFT offers several advantages as listed in Sec.~\ref{sec:introduction}, and leads to dramatic annotation cost reduction and computation efficiency. Besides, we have found that there are only seven fundamental patterns in CNN predictions, as summarized in Sec.~\ref{sec:selection_illustrated}. Multiple methods may be developed to select a particular pattern: entropy, Gaussian distance, and standard deviation would seek Pattern A, while diversity, variance, and divergence look for Pattern C. We are among the first to analyze the prediction patterns in active learning and investigate the effectiveness of typical patterns rather than comparing the many methods.

\begin{table}[t]
\footnotesize
\caption{
Active selection patterns analysis. We illustrate the relationships among seven prediction patterns and four active selection criteria, assuming that a candidate $\mathcal{C}_i$ has 11 augmented patches, and their probabilities $P_i$ are predicted by the current CNN, presented in the second column. With majority selection, the entropy and diversity are calculated based on the top 25\% (3 patches in this illustration) highest confidences on the dominant predicted category. The first choice of each method (column) is \textbf{bolded} and the second choice is \underline{underlined}.
}
\label{tab:predict_pattern}
\begin{center}
\includegraphics[width=1\linewidth]{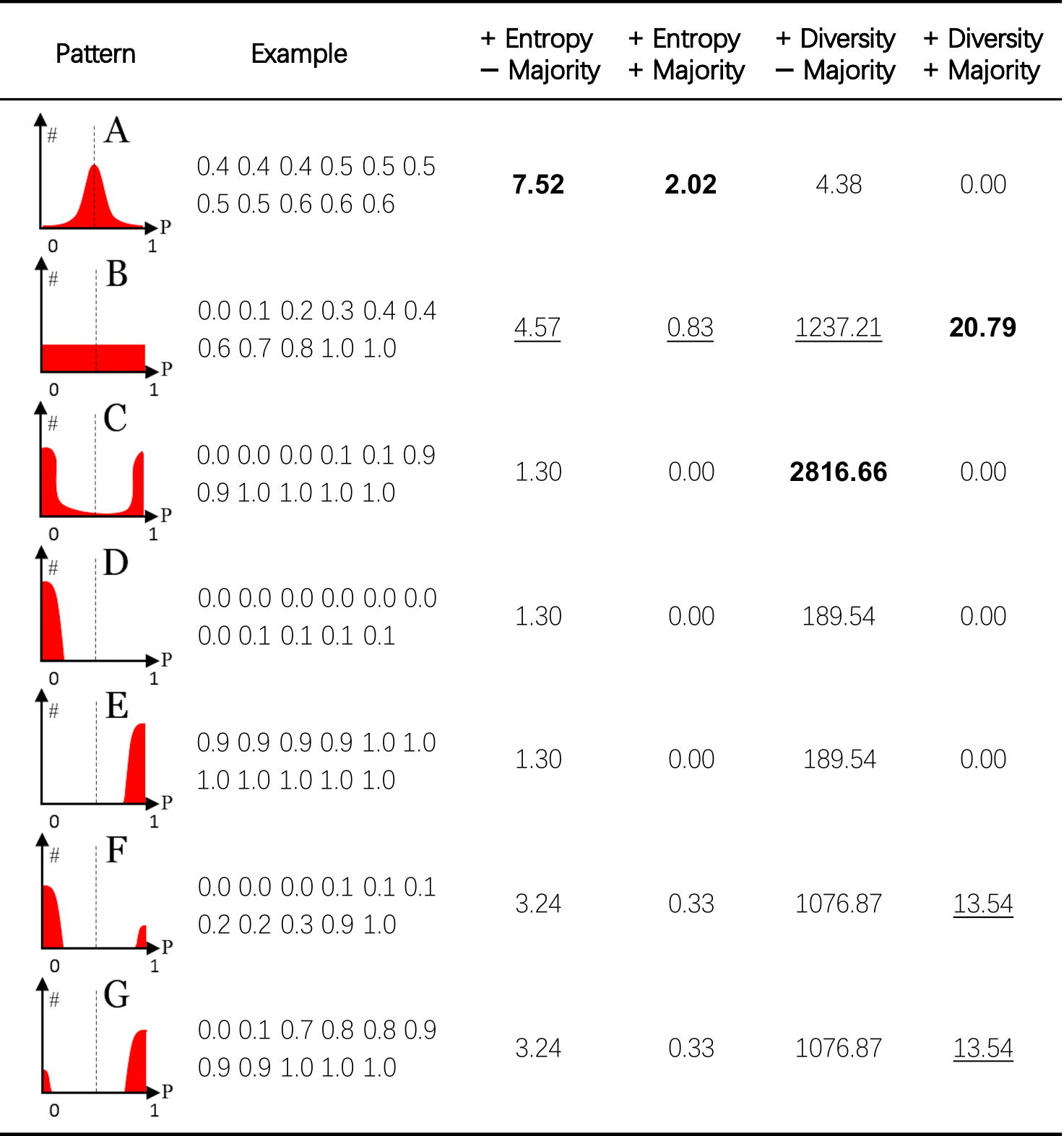}\\
\end{center}
\end{table}

\begin{algorithm*}[t] 
\small
\caption{ACFT -- Active, continual fine-tuning}
\label{alg:ACFT}
    \KwIn{\\
    \ \ \ $\mathcal{U}=\{\mathcal{C}_i\},$ $i\in [1,n]$ \{unlabeled pool $\mathcal{U}$\ contains $n$ candidates\}\\
    \ \ \ $\mathcal{C}_i=\{x_i^j\}$, $j\in [1,m]$ \{each $\mathcal{C}_i$ contains $m$ patches\}\\
    \ \ \ $M_0$: pre-trained CNN;
    \ $\alpha$: majority selection ratio;
    \ $b$: batch size;
    \ $\mathcal{Y}$: category set
    }
    \KwOut{\\
    \ \ \ $\mathcal{L}$: labeled candidates;
    \ $M_t$: fine-tuned CNN model at Step $t$
    }
    $\mathcal{L} \leftarrow \varnothing$;
    \ $t\leftarrow 1$\\
    \Repeat{classification performance in a validation set plateaus}{

    \For{ each $\mathcal{C}_i \in \mathcal{U}$ }
    {
        $P_i \leftarrow M_{t-1}(\mathcal{C}_i)$ \{outputs of $M_{t-1}$ given $\forall x \in \mathcal{C}_i$\}\\
        $\mathcal{C}'_i \leftarrow \mathcal{C}_i$ sorted in descending order according to the predicted dominant class $\hat{\textbf{y}}_i$ by Eq.~\ref{eq:dominate_class}, \ie $\hat{\textbf{y}}_i=\argmax_{y\in\mathcal{Y}} \frac{1}{m}\sum_{j=1}^{m}P^{j,y}_i$ \\
        $\mathcal{C}^{\alpha}_i \leftarrow$ top $\alpha\times 100\%$ of the patches of the sorted list $\mathcal{C}'_i$ \\
        Compute $\textbf{\em a}_i$ for $\mathcal{C}^{\alpha}_i$ by Eq.~\protect\ref{eq:R}, \ie $\textbf{\em a}_i =\lambda_1 \textbf{\em e}_i+\lambda_2 \textbf{\em d}_i\quad$\\
    }
    Sort $\mathcal{U}$ according to $\textbf{\em a}$ in descending order\\
    Compute sampling probability $\textbf{\em a}^{s}$ using sorted list $\textbf{\em a}'$ by Eq.~\ref{eq:randomness}, \ie $\textbf{\em a}'_i=(\textbf{\em a}'_i-\textbf{\em a}'_{\omega b})/(\textbf{\em a}'_1-\textbf{\em a}'_{\omega b}),\quad
    \textbf{\em a}^{s}_i=\textbf{\em a}'_i/\sum_i{\textbf{\em a}'_i},\quad \forall i \in [1,\omega b]$\\
    Associate labels for $b$ candidates with sampling probabilities: $\mathcal{Q}\leftarrow Q(\textbf{\em a}^{s},b)$ \\
    $P \leftarrow M_{t-1}(\mathcal{L})$ \{outputs of $M_{t-1}$ given $\forall x \in \mathcal{L}$\} \\
    Select misclassified candidates from $\mathcal{L}$ based on their annotation: $\mathcal{H} \leftarrow J(P, \mathcal{L})$ \\
    Fine-tune $M_{t-1}$ with $\mathcal{H}\bigcup\mathcal{Q}$: $M_t \leftarrow F(\mathcal{H}\bigcup\mathcal{Q},M_{t-1})$  \\
    $\mathcal{L} \leftarrow  \mathcal{L}\bigcup \mathcal{Q};  \quad \mathcal{U} \leftarrow \mathcal{U} \setminus \mathcal{Q};  \quad t\leftarrow t+1$ \\
    }
\end{algorithm*}

\begin{figure*}[t]
\begin{center}
\includegraphics[width=1.0\linewidth]{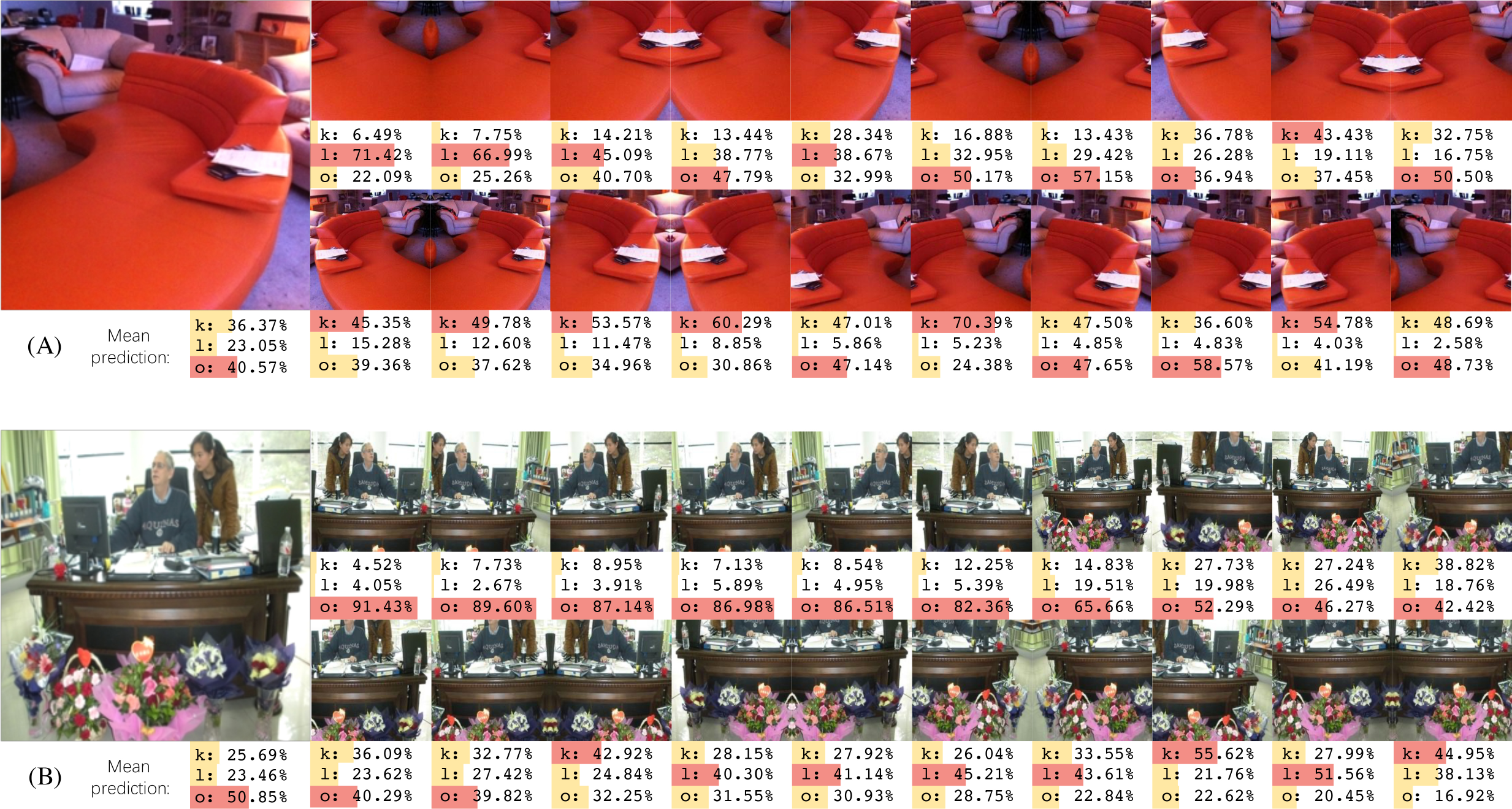}
\end{center}
\caption{
Automatic data augmentation inevitably generates noisy patches, and there is no need to classify all patches confidently. Therefore, we propose majority selection, which computes active selection criteria on only the top 25\% of the patches with the highest confidences on the dominant predicted category.
To demonstrate the necessity of majority selection, we illustrate two images (A and B) and their augmented patches, arranged according to the dominant category predicted by the CNN. 
Based on \textsc{Places-3}, Image A is labeled as {\em living room}, and its augmented patches are mostly incorrectly classified by the current CNN; therefore, including it in the training set is of great value. On the contrary, Image B is labeled as {\em office}, and the current CNN classifies most of its augmented patches as {\em office} with high confidence; labeling it would be of limited utility. 
Without majority selection, the criteria would mislead the selection, as it indicates that Image B is more diverse than Image A (297.52 vs. 262.39) while sharing similar entropy (17.33 vs. 18.50). With majority selection, the criteria show that Image A is considerably more uncertain and diverse than Image B, measured by either entropy (4.59 vs. 2.17) or diversity (9.32 vs. 0.35), and as expected, more worthy of labeling. From this active selection analysis, we remark that the majority selection is a critical component in our ACFT.
}
\label{fig:places_examples}
\end{figure*}

\begin{figure}[t]
\begin{center}
\includegraphics[width=1.0\linewidth]{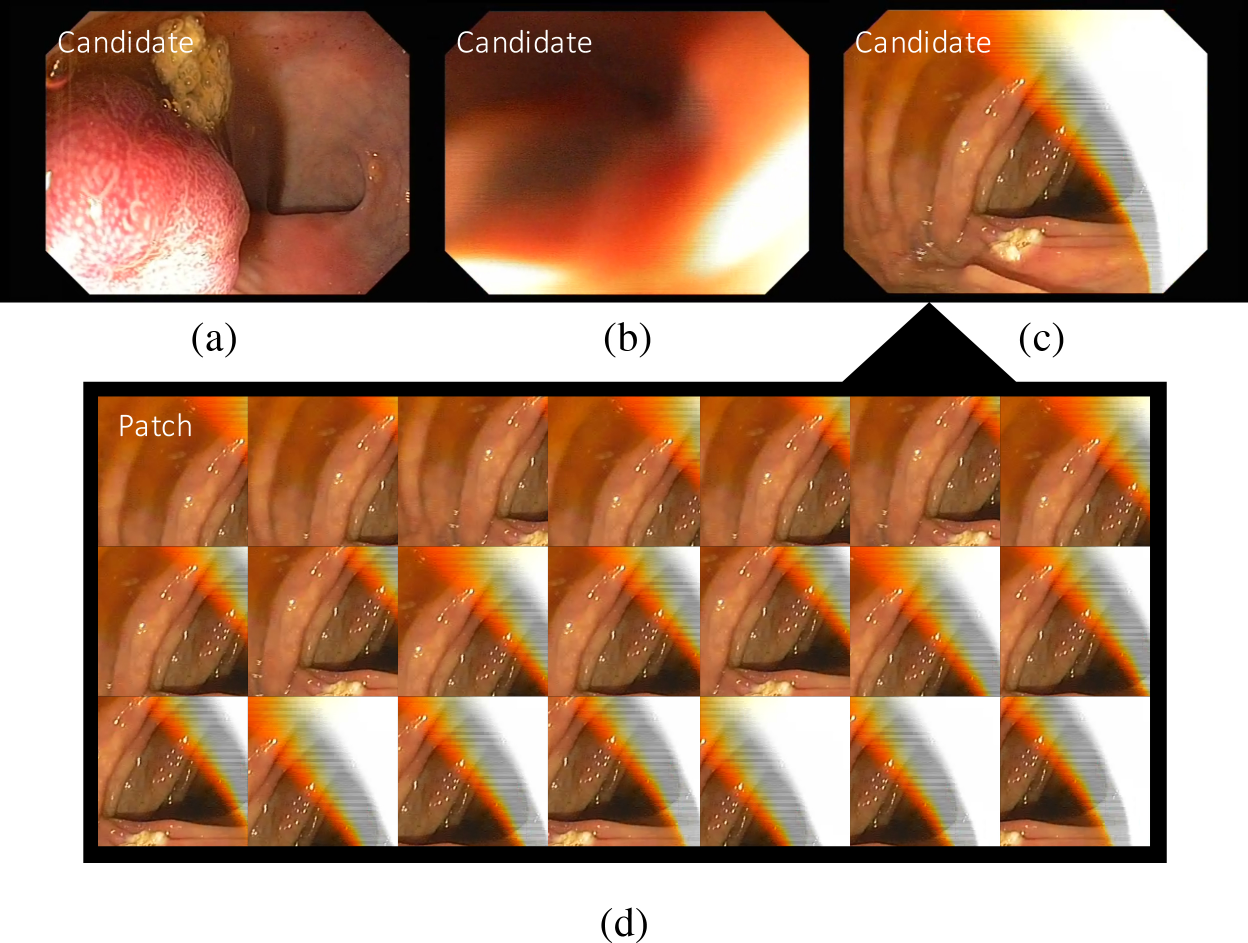}
\end{center}
\caption{
Three examples of colonoscopy frames: (a) informative, (b) non-informative, and (c) ambiguous. ``Ambiguous'' frames are labeled as ``informative" because experts label frames based on the overall quality: if over 75$\%$ of a frame (\ie candidate in this application) is clear,  the frame is considered ``informative''. As a result, an ambiguous candidate contains both clear and blurred components, and generates noisy labels at the patch level from automatic data augmentation. For example, the entire frame (c) is labeled as ``informative,'' but not all the patches (d) associated with this frame are ``informative", although they inherit the ``informative'' label. This limitation is the main motivation for the majority selection approach in our ACFT method.}
\label{fig:quality_assessment_dataset}
\end{figure}

\begin{figure}[t]
\centering
  \includegraphics[width=1.0\linewidth]{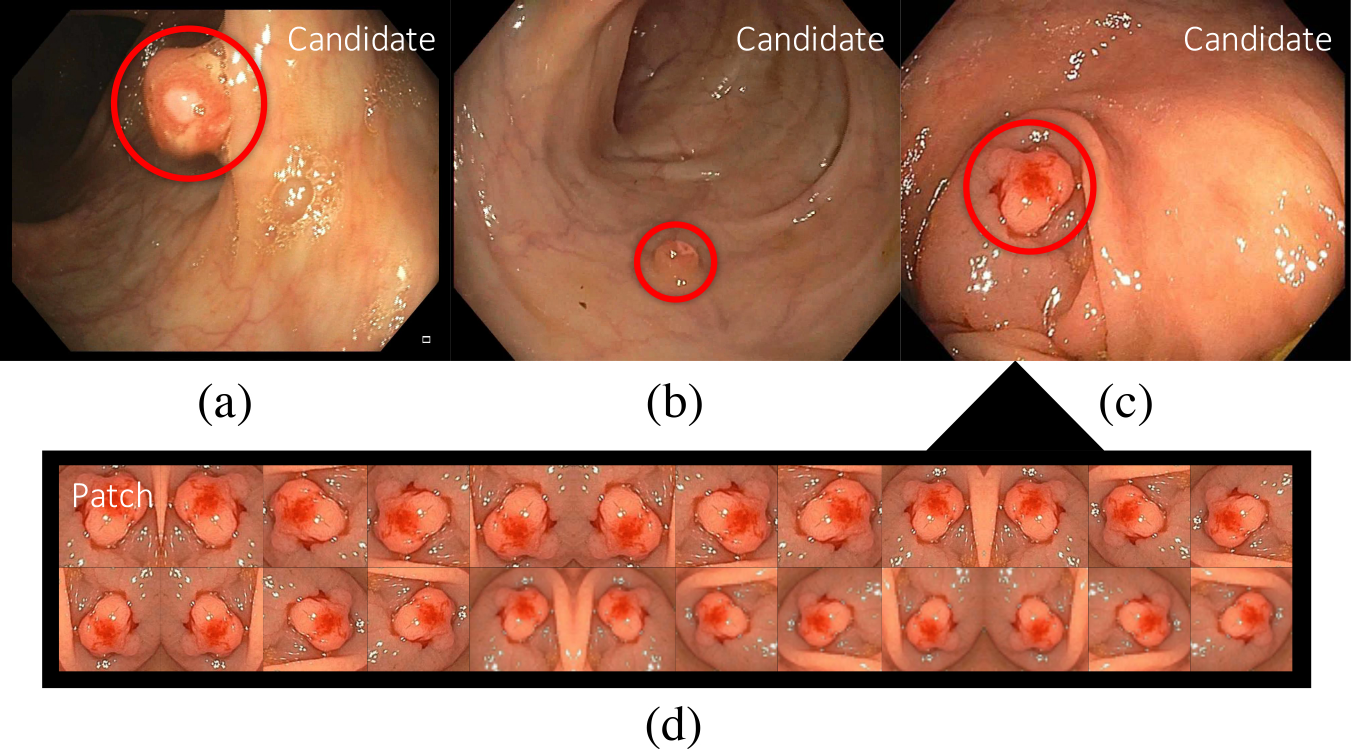}
\caption{
Polyps in colonoscopy videos with different shape and appearance.
}
\label{fig:polyp_detection_dataset}
\end{figure}

\begin{figure}[t]
\begin{center}
\includegraphics[width=1.0\linewidth]{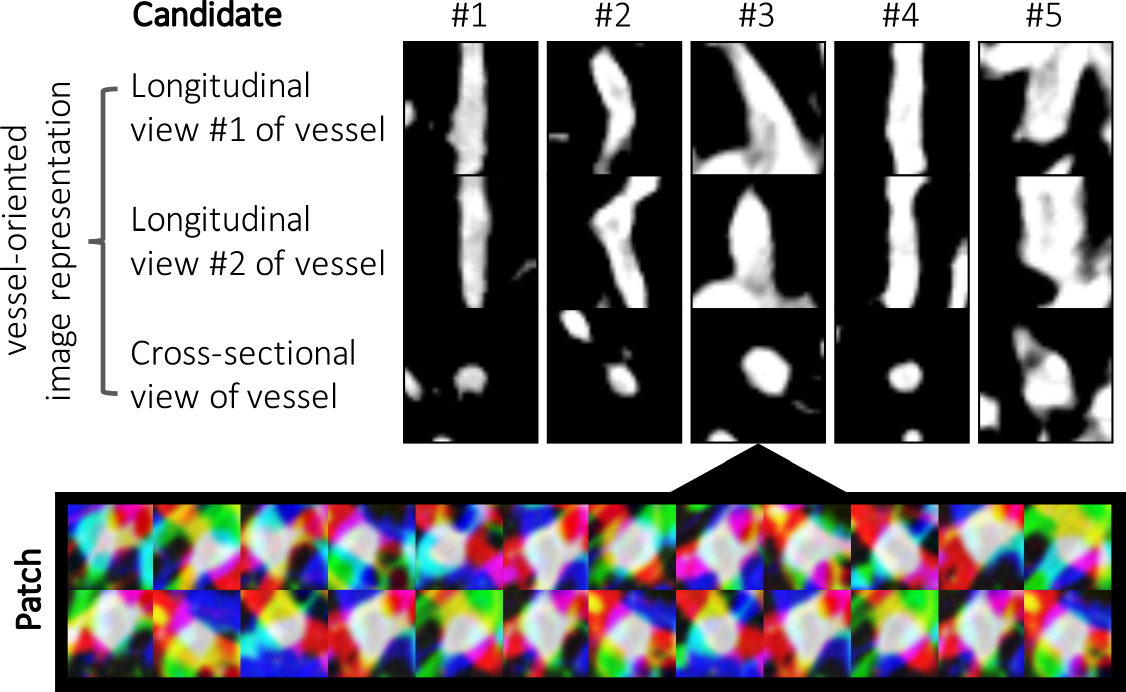}
\end{center}
\caption{
Five different pulmonary embolism candidates in the vessel-oriented image representation~\citep{tajbakhsh2015computer}. It was adopted in this work because it achieves great classification accuracy and accelerates CNN training convergence.
}
\label{fig:pulmonary_embolism_dataset}
\end{figure}

\section{Proposed method}
\label{sec:method}

ACFT was conceived in the context of computer-aided diagnosis (CAD) applied to medical imaging. A CAD system typically employs a candidate generator, which can quickly produce a set of candidates,  among which some are {\em true} positives and others are {\em false} positives. To train a classifier, each of the candidates must be labeled. In this work, an object to be labeled is considered as a ``candidate'' in general. We assume that each candidate takes one of $|\mathcal{Y}|$ possible labels.  To boost CNN performance for CAD systems, multiple patches are usually generated automatically for each candidate through data augmentation; those patches that are generated from the same candidate inherit the candidate's label. In other words, all labels are acquired at the candidate level. Mathematically, given a set of candidates, $\mathcal{U}=\{\mathcal{C}_1, \mathcal{C}_2, ..., \mathcal{C}_n\}$, where $n$ is the number of candidates, and each candidate $\mathcal{C}_i =\{x^{1}_i,x^{2}_i,...,x^{m}_i\}$ is associated with $m$ patches, our ACFT  algorithm iteratively selects a set of candidates for labeling as illustrated in~Alg.~\ref{alg:ACFT}.

ACFT is generic and applicable to many tasks in computer vision and image analysis. For clarity, we illustrate the ideas behind ACFT with the \textsc{Places-3} dataset~\citep{zhou2017places} for scene classification in natural images (see \figurename~\ref{fig:places_database}), where no candidate generator is needed, as each image may be directly regarded as a candidate.

Designing an active learning algorithm involves {\bf two key issues}: (1) how to determine the ``worthiness" of a candidate for annotation and (2) how to update the classifier/learner. In the following sections, we first illustrate our hypothesis in Sec.~\ref{sec:selection_illustrated} with \figurename~\ref{fig:places_examples} and \tableautorefname~\ref{tab:predict_pattern}, and then detail each of the components in our active selection criteria with its rationale and benefit.

\subsection{Illustrating active candidate selection}
\label{sec:selection_illustrated}

\figurename~\ref{fig:places_examples} shows the active candidate selection process for multi-class classification. To facilitate comprehension, \tableautorefname~\ref{tab:predict_pattern} illustrates the process in the context of binary classification.
Assuming the prediction of patch $x^{j}_i$ by the current CNN is $P_i^{j}$, we call the histogram of $P_i^{j}, j \in [1, m]$ the prediction pattern of candidate $\mathcal{C}_i$. As shown in Row 1 of Table~\ref{tab:predict_pattern}, in binary classification, there are seven typical prediction patterns:
\begin{enumerate}
\item Pattern A is mostly concentrated at 0.5, with a higher degree of uncertainty. Most active learning algorithms~\citep{settles2010active,guyon2011results} favor these types of candidates as they are effective for reducing uncertainty. 
\item Pattern B is flatter than Pattern A, as the patches' predictions are spread widely from 0 to 1 with a higher degree of inconsistency among the patches' predictions. Since all the patches belonging to a candidate are generated via data augmentation, they (at least the majority) are expected to make similar predictions. These types of candidates have the potential to significantly enhance the current CNN's performance.
\item Pattern C is clustered at the both ends, with a higher degree of diversity. These types of candidates are most likely associated with noisy labels at the patch level as illustrated in \figurename~\ref{fig:quality_assessment_dataset}(c), and they are the least favorable for use in active selection because they may cause confusion when fine-tuning the CNN.
\item Patterns D and E are clustered at either end (\ie 0 or 1), with a higher degree of certainty. These types of candidates should not undergo annotation at this step because it is likely the current CNN has correctly predicted them, and therefore these candidates would contribute very little towards fine-tuning the current CNN.
\item Patterns F and G have a higher degree of certainty for some of the patches' predictions but are associated with some outliers. These types of candidates are valuable because they are capable of smoothly improving the CNN's performance.  While such candidates might not make dramatic contributions, they do not significantly degrade the CNN's performance either.
\end{enumerate}

\subsection{Seeking worthy candidates}
\label{sec:entropy_diversity}

In active learning, the key is to develop criteria for determining candidate annotation ``worthiness''.
Our criteria for candidate ``worthiness'' are based on a simple, yet powerful, observation: all patches augmented from the same candidate (\figurename~\ref{fig:places_examples}) share the same label; therefore, they are expected to have similar predictions by the current CNN. As a result, their {\em entropy} and {\em diversity} provide a useful indicator of the ``power'' of a candidate for elevating the performance of the current CNN. Intuitively, entropy captures classification certainty---a higher uncertainty value denotes a greater degree of information (\eg pattern A in \tableautorefname~\ref{tab:predict_pattern}), whereas diversity indicates prediction consistency among the candidate patches---a higher diversity value denotes a greater degree of prediction inconsistency (\eg pattern C in \tableautorefname~\ref{tab:predict_pattern}). Formally, assuming that each candidate takes one of $|\mathcal{Y}|$ possible labels, we define the entropy and diversity of $\mathcal{C}_i$ as

\begin{equation}
\label{eq:active_criteria}
\begin{split}
&\textbf{\em e}_i = -\frac{1}{m}\sum_{k=1}^{|\mathcal{Y}|}{\sum_{j=1}^{m}{P_i^{j,k}\log{P_i^{j,k}}}}, \\
&\textbf{\em d}_i = \sum_{k=1}^{|\mathcal{Y}|}{\sum_{j=1}^m{\sum_{l=j}^m{(P_i^{j,k}-P_i^{l,k})\log{\frac{P_i^{j,k}}{P_i^{l,k}}}}}}
\end{split}
\end{equation}
Combining entropy and diversity yields
\begin{equation}
\label{eq:R}
\textbf{\em a}_i =\lambda_1 \textbf{\em e}_i+\lambda_2 \textbf{\em d}_i
\end{equation}
where $\lambda_1$ and $\lambda_2$ are trade-offs between entropy and diversity. We use two parameters for convenience,  to easily turn on/off entropy or diversity during experiments.

\subsection{Handling noisy labels via majority selection}
\label{sec:majority_selection}

Automatic data augmentation is essential for boosting CNN performance, but it inevitably generates ``hard'' samples for some candidates, as shown in \figurename~\ref{fig:quality_assessment_dataset}(c), injecting noisy labels. Therefore, to significantly enhance the robustness of our method, we compute entropy and diversity by selecting only a portion of the patches of each candidate according to the predictions by the current CNN.

Specifically, for each candidate $\mathcal{C}_i$ we first determine its dominant category, which is defined by the category with the highest confidence in the mean prediction. That is,
\begin{equation}
\label{eq:dominate_class}
\hat{\textbf{y}}_i =  \argmax_{y \in \mathcal{Y}} \frac{1}{m}\sum_{j=1}^{m}P^{j,y}_i
\end{equation}
where $P^{j,y}_i$ is the output of each patch $j$ from the current CNN given $\forall x \in \mathcal{C}_i$ on label $y$. After sorting $P_i$ according to dominant category $\hat{\textbf{y}}_i$, we apply Eq.~\ref{eq:R} to the top $\alpha\times$100$\%$ of the patches to construct the score matrix $\textbf{\em a}_i$ of size $\alpha m \times \alpha m$ for each candidate $\mathcal{C}_i$ in $\mathcal{U}$. Our proposed majority selection method automatically excludes the patches with noisy labels (see \tableautorefname~\ref{tab:predict_pattern}: diversity and diversity$^\alpha$) because of their low confidences.

\subsection{Injecting randomization in active selection}
\label{sec:randomness}

As discussed in~\citet{borisov2010active} and~\citet{ zhou2017fine}, simple random selection may outperform active selection at the beginning, because the active selection method depends on the current CNN selecting examples for labeling. As a result, a poor selection made at an early stage may adversely affect the quality of subsequent selections, whereas the random selection approach is less frequently locked into a poor hypothesis. In other words, the active selection method concentrates on exploiting the knowledge gained from the labels already acquired to further explore the decision boundary, whereas the random selection approach concentrates solely on exploration, and is thereby able to locate areas of the feature space where the classifier performs poorly. Therefore, an effective active learning strategy must strike a balance between exploration and exploitation.
Towards this end, we inject randomization into our method by selecting actively according to the sampling probability $\textbf{\em a}^{s}_i$.

\begin{equation}
\label{eq:randomness}
\begin{split}
\textbf{\em a}'_i=(\textbf{\em a}'_i-\textbf{\em a}'_{\omega b})/(\textbf{\em a}'_1-\textbf{\em a}'_{\omega b}), \\
 \textbf{\em a}^{s}_i=\textbf{\em a}'_i/\sum_i{\textbf{\em a}'_i},\quad \forall i \in [1,\omega b]
\end{split}
\end{equation}
where $\textbf{\em a}'_i$ is sorted $\textbf{\em a}_i$ according to its value in descending order, and $\omega$ is named random extension.
Suppose $b$ number of candidates are required for annotation. Instead of selecting top $b$ candidates, we extend the candidate selection pool to $\omega b$. Then we select candidates from this pool with their sampling probabilities $\textbf{\em a}^{s}_i$ to inject randomization.

\section{Experiments}
\label{sec:experiment}

\subsection{Medical applications}
\label{sec:medical_applications}

\subsubsection{Colonoscopy Frame Classification}

Image quality assessment in colonoscopy can be viewed as an image classification task whereby an input image is labeled as either \textit{informative} or \textit{non-informative}. 
One way to measure the quality of a colonoscopy procedure is to monitor the quality of the captured images. Such quality assessment can be used during live procedures to limit low-quality examinations or, in a post-processing setting, for quality monitoring purposes.
In this application, colonoscopy frames are regarded as \textit{candidates}, since the labels (informative or non-informative) are associated with frames as illustrated in \figurename~\ref{fig:quality_assessment_dataset}(a---c).  In total, there are 4,000 colonoscopy candidates from 6 complete colonoscopy videos. A trained expert then manually labeled the collected images as informative or non-informative (line 11 in Alg.~\ref{alg:ACFT}). A gastroenterologist further reviewed the labeled images for corrections. The labeled frames are separated at the video level into training and test sets, each containing approximately 2,000 colonoscopy frames. For data augmentation, we extracted 21 patches from each frame as shown in \figurename~\ref{fig:quality_assessment_dataset}(d).

\subsubsection{Polyp Detection}

Polyps, as shown in \figurename~\ref{fig:polyp_detection_dataset}, can present themselves in the colonoscopy with substantial variations in color, shape, and size. The variable appearance of polyps can often lead to misdetection, particularly during long and back-to-back colonoscopy procedures where fatigue negatively affects the performance of colonoscopists. Computer-aided polyp detection may enhance optical colonoscopy screening accuracy by reducing polyp misdetection. 
In this application, each polyp detection is regarded as a \textit{candidate}.
The dataset contains 38 patients with one video each. The training dataset is composed of 21 videos (11 with polyps and 10 without polyps), while the testing dataset is composed of 17 videos (8 videos with polyps and 9 videos without polyps). At the video level, the candidates are divided into the training dataset (16,300 candidates) and test dataset (11,950 candidates). At each polyp candidate location with the given bounding box, we performed data augmentation by a factor $f\in \{1.0,1.2,1.5\}$. At each scale, we extracted patches after the candidate is translated by 10 percent of the resized bounding box in vertical and horizontal directions. We further rotated each resulting patch 8 times by mirroring and flipping. The patches generated by data augmentation belong to the same candidate. Each candidate contains 24 patches.

\subsubsection{Pulmonary Embolism Detection}

Pulmonary embolism (PE) is a major national health problem, and computer-aided PE detection could play a major role in improving PE diagnosis and decreasing the reading time required for CTPA datasets.
We employed a database consisting of 121 CTPA datasets with a total of 326 PE instances. Each PE detection is regarded as a \textit{candidate} with 50 patches. We divided candidates at the patient level into a training dataset, with 434 true positives (199 unique PE instances) and 3,406 false positives, and a testing dataset, with 253 true positives (127 unique PE instances) and 2,162 false positives. The overall PE probability is calculated by averaging the probabilistic prediction generated for the patches within a given PE candidate after data augmentation.

\begin{table}[t]
\footnotesize
\begin{center}
\begin{threeparttable}

\caption{Active learning strategy definition. We have codified different learning strategies covering the makeup of training samples and the initial CNN weights of fine-tuning.}
\label{tab:terminology}
    \begin{tabular}{p{0.06\textwidth}p{0.36\textwidth}}
    \hline
    \textbf{Code} & \textbf{Description of learning strategy} \\
    \hline
    RFT$_{(LQ)}$ & {\scriptsize Fine-tuning from $M_{0}$ using $\mathcal{L}$ and randomly selected $\mathcal{Q}$} \\
    AFT$_{(LQ)}$ & {\scriptsize Fine-tuning from $M_0$ using $\mathcal{L}$ and actively selected $\mathcal{Q}$} \\
    ACFT$_{(Q)}$ & {\scriptsize Continual fine-tuning from $M_{t-1}$ using actively selected $\mathcal{Q}$ only} \\
    ACFT$_{(LQ)}$ & {\scriptsize Continual fine-tuning from $M_{t-1}$ using $\mathcal{L}$ and actively selected $\mathcal{Q}$} \\
    ACFT$_{(HQ)}$ & {\scriptsize Continual fine-tuning from $M_{t-1}$ using $\mathcal{H}$ and actively selected $\mathcal{Q}$} \\
    \hline
    \end{tabular}
    \begin{tablenotes}
        \footnotesize
        \item[1] $\mathcal{L}$: Annotated candidates.
        \item[2] $\mathcal{Q}$: Newly annotated candidates.
        \item[3] $\mathcal{H}$: Misclassified candidates.
        \item[4] $M_0$: Pre-trained CNNs from large scale dataset (like \textsc{ImageNet}).
        \item[5] $M_{t-1}$: Pre-trained CNNs from last active selecting step.
    \end{tablenotes}
\end{threeparttable}
\end{center}
\end{table}

\begin{table}[t]
\footnotesize 
\begin{center}
\begin{threeparttable}
\caption{Learning parameters used for training and fine-tuning of AlexNet for AFT in our experiments. $\mu$ is the momentum, $lr_{fc8}$ is the learning rate of the weights in the last layer, $\alpha$ is the learning rate of the weights in the rest layers, and $\gamma$ determines how $lr$ decreases over epochs. ``Epochs" indicates the number of epochs used in each step. For ACFT, all the parameters are set to the same as AFT except the learning rate $lr$, which is set to 1/10 of that for AFT.}
\label{tab:hyperparameter}
\begin{tabular}{p{0.21\textwidth}P{0.015\textwidth}P{0.03\textwidth}P{0.03\textwidth}P{0.02\textwidth}P{0.035\textwidth}}
\hline
\textbf{Applications} & \boldsymbol{$\mu$} & \boldsymbol{$lr$} & \boldsymbol{$lr_{fc8}$} & \boldsymbol{$\gamma$} & \textbf{epoch} \\
\hline
Colonoscopy frame classification & 0.9 & 1e-4 & 1e-3 & 0.95 & 8 \\
Polyp detection & 0.9 & 1e-4 & 1e-3 & 0.95 & 10 \\
Pulmonary embolism detection & 0.9 & 1e-3 & 1e-2 & 0.95 & 5 \\
\hline
\end{tabular}
\end{threeparttable}
\end{center}
\end{table}

\begin{figure*}[t]
\begin{center}
\includegraphics[width=1.0\linewidth]{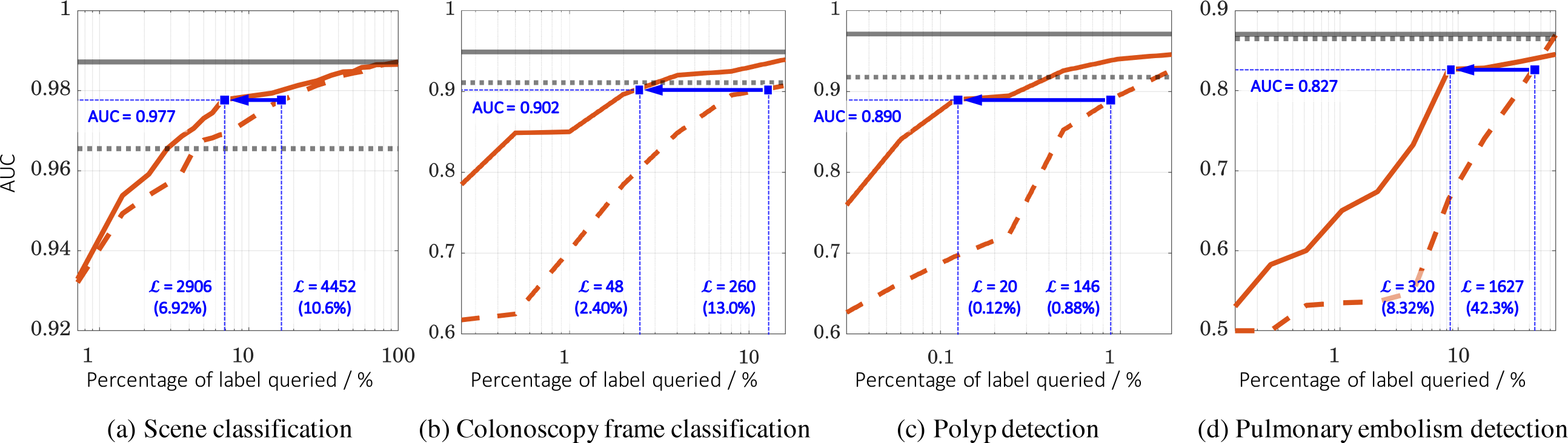}
\end{center}
\caption{
ACFT aims to minimize the number of samples for experts to label by iteratively recommending the most informative and representative samples. For scene classification (a), by actively selecting 2,906 images (6.92$\%$ of the entire dataset), ACFT (solid orange) can offer equivalent performance to the use of 4,452 images through random selection, thus saving 34.7$\%$ annotation cost relative to random fine-tuning (RFT in dashed orange). Furthermore, with 1,176 actively-selected images (2.80$\%$ of the whole dataset), ACFT can achieve performance equivalent to full training (dashed black) using 42,000 images, thereby saving 97.2$\%$ annotation cost (relative to full training). In (b)---(d), we highlight the major results that compared with RFT, our ACFT can reduce the cost of annotation by 81.5$\%$ for colonoscopy frame classification, 86.3$\%$ for polyp detection, and 80.3$\%$ for pulmonary embolism detection. Following the standard active learning experimental setup, both ACFT and RFT select samples from the remaining training dataset; they will eventually use the same whole training dataset, naturally yielding similar performance at the end. However, the goal of active learning is to find such sweet spots where a learner can achieve an acceptable performance using the least number of labeled samples.
}
\label{fig:overall_result}
\end{figure*}

\begin{figure*}[t]
\begin{center}
\includegraphics[width=1\linewidth]{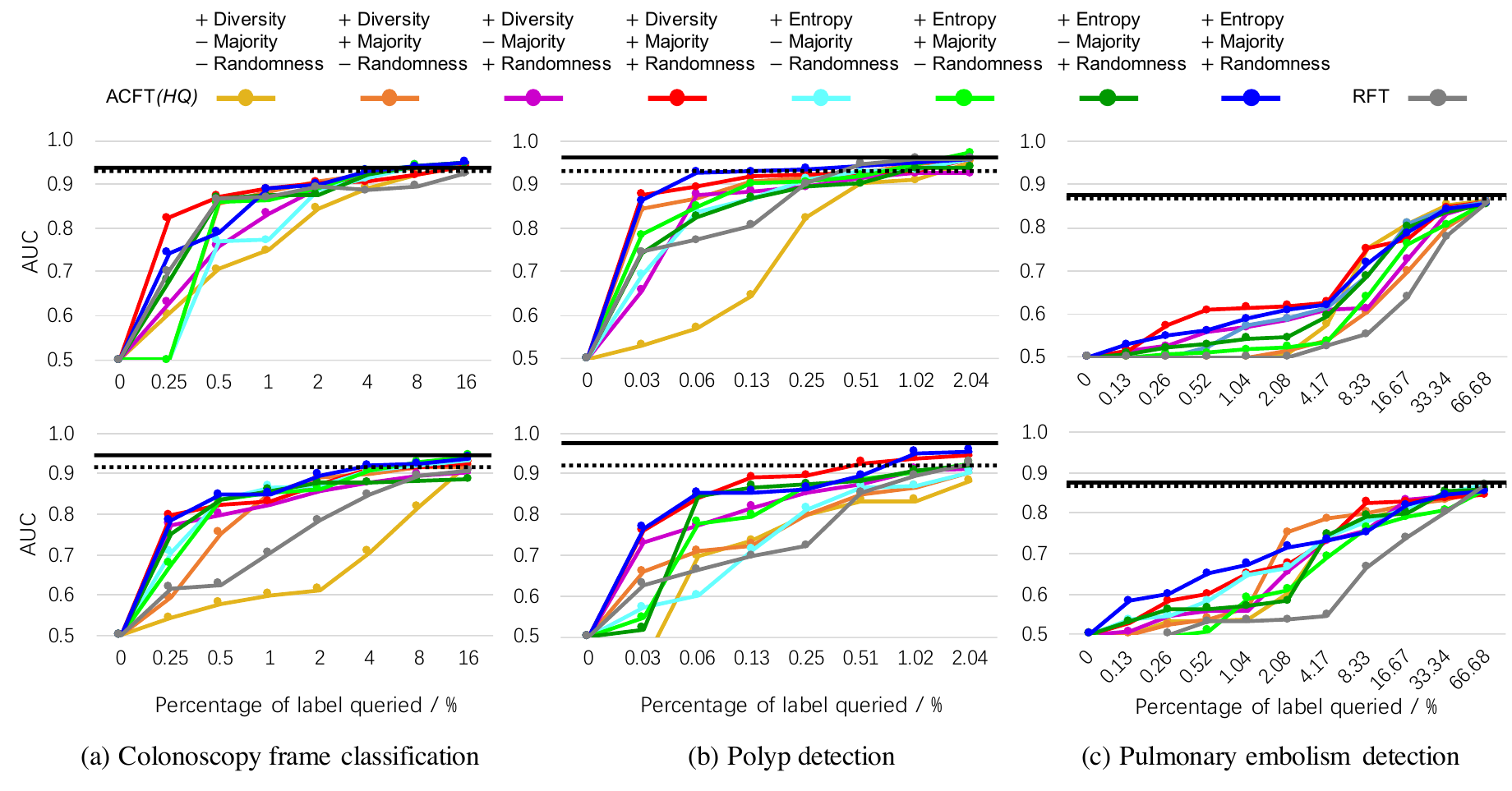}
\end{center}
\caption{Comparing eight active selection approaches with random selection on AlexNet~\citep{krizhevsky2012imagenet} (top panel) and GoogLeNet~\citep{szegedy2015going} (bottom panel) for our three distinct medical applications, including (a) colonoscopy frame classification, (b) polyp detection, and (c) pulmonary embolism detection, demonstrates consistent patterns with AlexNet. The solid black line denotes the current state-of-the-art performance of fine-tuning using full training data and the dashed black line denotes the performance of training from scratch using full training data.}
\label{fig:selection_approaches_comparison}
\end{figure*}

\begin{figure*}[t]
\footnotesize
\begin{center}
  \includegraphics[width=1.0\linewidth]{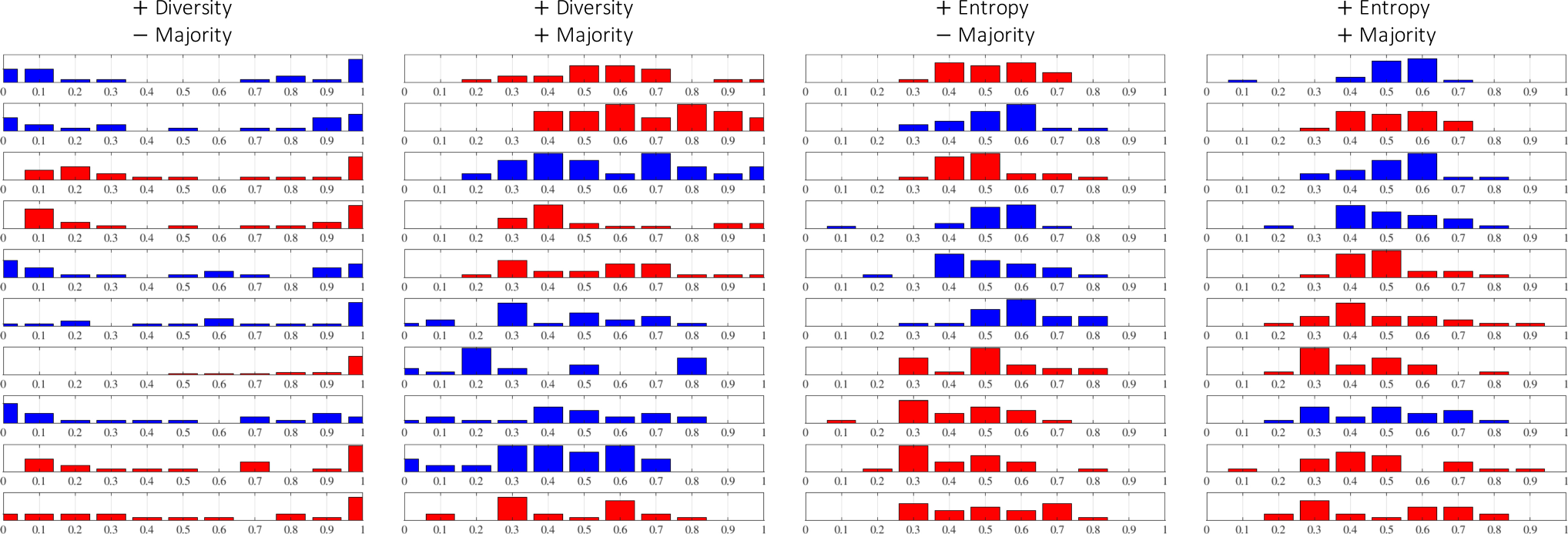}
\end{center}
  \caption{
  Distribution of predictions for the top ten candidates actively selected by the four ACFT methods at Step 3 in colonoscopy frame classification. Positive candidates are shown in red and negative candidates are shown in blue. This visualization confirms the assumption in \tableautorefname~\ref{tab:predict_pattern} that diversity+majority selection criteria prefers Pattern B whereas diversity suggests Pattern C; both entropy and entropy+majority favor Pattern A due to its higher degree of uncertainty. However, in this case at Step 3, with entropy+majority selection criteria, there are no more candidates with Pattern A; therefore, candidates with Pattern B are selected.
  }
\label{fig:predicted_distribution}
\end{figure*}

\begin{table*}[t]
\scriptsize
\begin{center}
\begin{threeparttable}

    \caption{Comparison of proposed active learning strategies and selection criteria. As measured by the Area under the Learning Curve (ALC), bolded values in the table indicate the outstanding learning strategies (see \tableautorefname~\ref{tab:terminology}) using certain active selection criteria, and starred values represent the best performance taking both learning strategies and active selection criteria into consideration. For all three applications, we report baseline performance of random fine-tuning (RFT) using AlexNet in the table footnote. Considering the variance of random sampling for each active learning step, we conduct five independent trials for RFT and report the mean and standard deviation (mean$\pm$s.d.).}
    \label{tab:main_results}
    \begin{tabular}{p{0.07\textwidth}p{0.055\textwidth}|p{0.08\textwidth}p{0.08\textwidth}p{0.08\textwidth}p{0.08\textwidth}|p{0.08\textwidth}p{0.08\textwidth}p{0.08\textwidth}p{0.08\textwidth}}
    \hline
    Application & Learning strategy & $+$ Diversity\newline $-$ Majority\newline $-$ Randomness & $+$ Diversity\newline $+$ Majority\newline $-$ Randomness & $+$ Diversity\newline $-$ Majority\newline $+$ Randomness & $+$ Diversity\newline $+$ Majority\newline $+$ Randomness & $+$ Entropy\newline $-$ Majority\newline $-$ Randomness & $+$ Entropy\newline $+$ Majority\newline $-$ Randomness & $+$ Entropy\newline $-$ Majority\newline $+$ Randomness & $+$ Entropy\newline $+$ Majority\newline $+$ Randomness \\
    \hline
    \multirow{4}{*}{\tabincell{l}{Colonoscopy\\frame\\classification}} & ACFT$_{(Q)}$ & 0.8375 & 0.8773 & 0.8995 & 0.9160 & 0.8444 & 0.8227 & 0.9136 & 0.9061\\
    & ACFT$_{(LQ)}$ & 0.8501 & 0.8956 & 0.9083 & 0.9262 & 0.9149 & 0.9051 & 0.9033 & 0.9223\\
    & AFT$_{(LQ)}$ & \textbf{0.9183} & \textbf{0.9253} & \textbf{0.9299} & \textbf{0.9344}$^\star$ & \textbf{0.9219} & 0.9180 & \textbf{0.9268} & 0.9291\\
    & ACFT$_{(HQ)}$ & 0.9048 & 0.9236 & 0.9241 & 0.9179 & 0.9198 & \textbf{0.9266} & 0.9257 & \textbf{0.9293}\\
    \hline
    \multirow{4}{*}{\tabincell{l}{Polyp\\detection}} & ACFT$_{(Q)}$ & 0.8669 & 0.9023 & 0.8984 & 0.9168 & 0.8834 & 0.8656 & 0.9034 & 0.9271\\
    & ACFT$_{(LQ)}$ & 0.9195 & 0.9142 & \textbf{0.9497} & \textbf{0.9488} & 0.9204 & 0.9255 & \textbf{0.9475} & 0.9444\\
    & AFT$_{(LQ)}$ & \textbf{0.9242} & 0.9285 & 0.9353 & 0.9355 & 0.9292 & 0.9238 & 0.9367 & \textbf{0.9522}$^\star$ \\
    & ACFT$_{(HQ)}$ & 0.9013 & \textbf{0.9370} & 0.9116 & 0.9363 & \textbf{0.9321} & \textbf{0.9436} & 0.9196 & 0.9443\\
    \hline
    \multirow{4}{*}{\tabincell{l}{Pulmonary\\embolism\\detection}} & ACFT$_{(Q)}$ & 0.7828 & 0.7911 & 0.7690 & 0.7977 & 0.7855 & 0.7736 & 0.7296 & 0.7833\\
    & ACFT$_{(LQ)}$ & 0.8083 & \textbf{0.8176} & 0.7975 & \textbf{0.8263} & 0.8032 & \textbf{0.8086} & 0.8022 & \textbf{0.8245}\\
    & AFT$_{(LQ)}$ & 0.7650 & 0.7973 & 0.7978 & 0.8040 & 0.7917 & 0.7878 & 0.7964 & 0.8222\\
    & ACFT$_{(HQ)}$ & \textbf{0.8272}$^\star$ & 0.7876 & \textbf{0.8047} & 0.8245 & \textbf{0.8218} & 0.7995 & \textbf{0.8155} & 0.8205\\
    \hline
    \end{tabular}
    \begin{tablenotes}
        \footnotesize
        \item[1] RFT in colonoscopy frame classification: ALC = 0.8958$\pm$0.0176
        \item[2] RFT in polyp detection: ALC = 0.9358$\pm$0.0130
        \item[3] RFT in pulmonary embolism detection: ALC = 0.7849$\pm$0.0261
    \end{tablenotes}
\end{threeparttable}
\end{center}
\end{table*}

\subsection{Baselines and implementation}

\subsubsection{Active learning strategy baselines}
\label{sec:active_learning_strategy_baselines}

\citet{tajbakhsh2016convolutional} reported the state-of-the-art performance of fine-tuning and learning from scratch using entire datasets, which are used to establish baseline performance for comparison. These authors also investigated the performance of (partial) fine-tuning using a sequence of partial training datasets, but our dataset partitions are different from theirs. Therefore, for fair comparison with their approach, we introduce RFT, which fine-tunes the original CNN model $M_0$  from the beginning, using all available labeled samples $\mathcal{L}\bigcup\mathcal{Q}$, where $\mathcal{Q}$ is randomly selected at each step. 

We summarized several active learning strategies in \tableautorefname~\ref{tab:terminology}. Studying different active learning strategies is important because active learning procedure can be very computationally inefficient in practice, in terms of label reuse and model reuse. We present two strategies that aim at overcoming the above limitations. First, we propose to combine newly annotated data with the labeled data that is misclassified by the current CNN. Second, we propose continual fine-tuning to speed up model training and, in turn, encourage data reuse.
ACFT$_{(HQ)}$ denotes the optimized learning strategy, which continually fine-tunes the current CNN model $M_{t-1}$ using newly annotated candidates enlarged by those misclassified candidates; that is, $\mathcal{Q}\bigcup\mathcal{H}$. 
Compared with other learning strategy baselines~\citep{tajbakhsh2016convolutional, zhou2017fine,zhou2019integrating} as codified in \tableautorefname~\ref{tab:terminology}, ACFT$_{(HQ)}$ saves training time through faster convergence compared with repeatedly fine-tuning the original pre-trained CNN, and boosts performance by eliminating easy samples, focusing on hard samples, and preventing catastrophic forgetting. In all three applications, our ACFT begins with an empty training dataset and directly uses pre-trained CNNs (AlexNet and GoogLeNet) on ImageNet.

\subsubsection{Experimental settings}
We have investigated the effectiveness of ACFT in four applications: scene classification, colonoscopy frame classification, polyp detection, and pulmonary embolism (PE) detection. Ablation studies have been conducted to confirm the significant design of our majority selection and randomization, built upon conventional entropy and diversity based active selection criteria.
For all four applications, we set $\alpha$ to 1/4 and $\omega$ to 5.
The deep learning library Matlab and Caffe are utilized to implement active learning and transfer learning (more details can be found at \href{https://github.com/MrGiovanni/Active-Learning}{https://github.com/MrGiovanni/Active-Learning}). We based our experiments on AlexNet and GoogLeNet because their architectures offer an optimal depth balance, deep enough to investigate the impact of ACFT and AFT on pre-trained CNN performance, but shallow enough to conduct experiments quickly. The learning parameters used for training and fine-tuning of AlexNet in our experiments are summarized in \tableautorefname~\ref{tab:hyperparameter}. The Adam optimizer is utilized to optimize the objective functions described in our paper. The batch size is 512 in the learning procedure.

\section{Results}
\label{sec:result}

In this section, \figurename~\ref{fig:overall_result} begins with an overall performance between our active continual fine-tuning (ACFT) and random fine-tuning (RFT), revealing the amount of annotation effort that has been reduced in each application. \figurename~\ref{fig:selection_approaches_comparison} compares eight different active selection criteria, demonstrating that majority selection and randomness are critical in finding the most representative samples to elevate the current CNN's performance.
\figurename~\ref{fig:predicted_distribution} further presents the observed distribution from each active selection criteria, qualitatively confirming the rationale of our devised candidate selecting approaches.
\tableautorefname~\ref{tab:main_results} finally compares four different active learning strategies, suggesting that continual fine-tuning using newly annotated candidates enlarged by those misclassified candidates significantly saves computational resources while maintaining the compelling performance in all three medical applications.

\subsection{ACFT reduces 35\% annotation effort in scene classification}

\figurename~\ref{fig:overall_result}(a) compares ACFT with RFT in scene classification using the \textsc{Places-3} dataset. For RFT, six different sequences are generated via systematic random sampling. The final curve is plotted showing the average performance of six runs. As shown in \figurename~\ref{fig:overall_result}(a), ACFT, with only 2,906 candidate queries, can achieve performance equivalent to RFT with 4,452 candidate queries, as measured by the Area Under the Curve (AUC); moreover, using only 1,176 candidate queries, ACFT can achieve performance equivalent to full training using all 42,000 candidates. Therefore, 34.7$\%$ of the RFT labeling costs and 97.2$\%$ of the full training costs could be saved using ACFT. When nearly 100$\%$ training data are used, the performance continues to improve, suggesting that the dataset size is still insufficient, given 22 layers GoogLeNet architecture. ACFT is a general algorithm that is not only useful for medical datasets but other datasets as well, and is also effective for multi-class problems.

\subsection{ACFT reduces 82\% annotation effort in colonoscopy frame classification}
\label{sec:result_colonoscopy_frame_classification}

\figurename~\ref{fig:overall_result}(b) shows that ACFT, with approximately 120 candidate queries (6$\%$), achieves performance equivalent to a 100$\%$ trained dataset fine-tuned from AlexNet (solid black line, AUC = 0.9366), and, with only 80 candidate queries (4$\%$), can achieve performance equivalent to a 100$\%$ training dataset learned from scratch (dashed black line, AUC = 0.9204). Using only 48 candidate queries, ACFT equals the performance of RFT at 260 candidate queries. Therefore, about 81.5$\%$ of the labeling cost associated with with RFT in colonoscopy frame classification is recovered using ACFT. Detailed analysis in \figurename~\ref{fig:selection_approaches_comparison} reveals that during the early stages, RFT yields performance superior to some of the active selecting processes because: 
1) random selection gives samples with the positive-negative ratio compatible with the testing and validation dataset; 2) the pre-trained CNN gives poor predictions in the domain of medical imaging, as it was trained by natural images. Its output probabilities are mostly inconclusive or even opposite, yielding poor selection scores. However, with randomness injected, as described in Sec.~\ref{sec:randomness}, ACFT (+majority and +randomness) shows superior performance, even at early stages, with continued performance improvement during subsequent steps (see the red and blue curves in \figurename~\ref{fig:selection_approaches_comparison}). Besides, evidenced by \tableautorefname~\ref{tab:main_results}, ACFT performs comparably with AFT, but, unlike the latter, does not require use of the entire labeled dataset or fine-tuning from the beginning. 

\subsection{ACFT reduces 86\% annotation effort in polyp detection}

\figurename~\ref{fig:overall_result}(c) shows that ACFT, with approximately 320 candidate queries (2.04$\%$), can achieve performance equivalent to a 100$\%$ training dataset fine-tuned from AlexNet (solid black line, AUC = 0.9615), and, with only 10 candidate queries (0.06$\%$), can achieve  performance equivalent to a 100$\%$ training dataset learned from scratch (dashed black line, AUC = 0.9358). Furthermore, ACFT, using only 20 candidate queries, achieves performance equivalent to RFT using 146 candidate queries. Therefore, nearly 86.3$\%$ of the labeling cost associated with the use of RFT for polyp detection could be recovered with our method. The fast convergence and outstanding performance of ACFT is attributable to the majority selection and randomization method, which can both efficiently select the informative and representative candidates while excluding those with noisy labels, yet still boost the performance during the early stages. For example, the diversity criteria, if without using majority selection, would strongly favor candidates whose prediction pattern resembles Pattern C (see \tableautorefname~\ref{tab:predict_pattern}), thus performing poorer than RFT due to noisy labels generated through data augmentation.

\subsection{ACFT reduces 80\% annotation effort in pulmonary embolism detection}

\figurename~\ref{fig:overall_result}(d) shows that ACFT, with 2,560 candidate queries (66.68$\%$) nearly achieves performance equivalent to both the 100$\%$ training dataset fine-tuned from AlexNet and learning from scratch (solid black line and dashed black line, where AUC = 0.8763 and AUC = 0.8706, respectively). With 320 candidate queries, ACFT can achieve the performance equivalent to RFT using 1,627 candidate queries. Based on this analysis, the cost of annotation in pulmonary embolism detection can be reduced by 80.3\% using ACFT compared with RFT.

\subsection{Observations on active selection criteria}
\label{sec:observation_criteria}

We meticulously monitored the active selection process and examined the selected candidates. For example, we include the top ten candidates selected by the four ACFT methods at Step 3 in colonoscopy frame classification in \figurename~\ref{fig:predicted_distribution}. From this process, we have observed the following:
\begin{itemize}[noitemsep]
\item Patterns A and B are dominant in the earlier stages of ACFT as the CNN has not been fine-tuned properly to the target domain;
\item Patterns C, D and E are dominant in the later stages of ACFT as the CNN has been largely fine-tuned on the target dataset;
\item Majority selection is effective for excluding Patterns C, D, and E, whereas entropy only (without the majority selection) can handle Patterns C, D, and E reasonably well;
\item Patterns B, F, and G generally make good contributions to elevating the current CNN's performance;
\item Entropy and entropy+majority favor Pattern A due to its higher degree of uncertainty, and;
\item Diversity+majority prefers Pattern B whereas diversity prefers Pattern C. This is why diversity may cause sudden disturbances in the CNN's performance and why diversity+majority is generally preferred.

\end{itemize}

\subsection{Comparison of proposed learning strategies}
\label{sec:comparison_learning_strategies}

As summarized in \tableautorefname~\ref{tab:terminology}, several active learning strategies can be derived. The prediction performance was evaluated according to the Area under the Learning Curve (ALC), in which the learning curve plots AUC as a function of the number of labels queried~\citep{guyon2011results}, computed on the testing dataset. \tableautorefname~\ref{tab:main_results} shows the ALC of ACFT$_{(Q)}$, ACFT$_{(LQ)}$, AFT$_{(LQ)}$ and ACFT$_{(HQ)}$ compared with RFT. Our comprehensive experiments have demonstrated that:

\begin{enumerate}
    \item ACFT$_{(Q)}$ considers only newly selected candidates for fine-tuning, resulting in an unstable CNN performance due to the catastrophic forgetting of the previous samples;
    \item ACFT$_{(LQ)}$ requires a careful parameter adjustment. Although its performance is acceptable, it requires the same computing time as AFT$_{(LQ)}$, indicating that there is no advantage to continually fine-tuning the current CNN;
    \item AFT$_{(LQ)}$ shows the most reliable performance compared with ACFT$_{(Q)}$ and ACFT$_{(LQ)}$;
    \item The optimized version, ACFT$_{(HQ)}$, shows comparable performance to AFT$_{(LQ)}$ and occasionally outperforms AFT$_{(LQ)}$ by eliminating easy samples, focusing on hard samples, and preventing catastrophic forgetting.
\end{enumerate}

In summary, our results suggest that (1) it is unnecessary to re-train models repeatedly from scratch for each active learning step and (2) learning newly annotated candidates plus a small portion of the misclassified candidates leads to equivalent performance to using the entire labeled set.

\section{Discussion}
\label{sec:discussion}

\subsection{How does intra-diversity differ from inter-diversity?}

Since measuring diversity between selected samples and unlabeled samples is computationally intractable, especially for a large pool of data~\citep{sourati2016classification}, the existing diversity sampling cannot be applied directly to our real-world medical applications. To name a few, selection criteria $R$ in \citet{chakraborty2015active} involves all unlabeled samples (patches). There are 391,200 training patches for polyp detection, and computing their $R$ would demand 1.1 TB memory (391,00$^2\times$8). In addition, their algorithms for batch selection are based on the truncated power method~\citep{yuan2013truncated}, which is unable to find a solution even for our smallest application (colonoscopy frame classification with 42,000 training patches). \citet{holub2008entropy} cannot be directly used for our real-world applications either, as it has a complexity of $\mathcal{O}(L^3\times N^3)$ and requires to train $L\times N$ classifiers in each step, where $N$ indicates the number of unlabeled patches and $L$ indicates the number of classes. In addressing the computational complexity problem, we exploit the inherent consistency among the patches that are augmented from the same sample, making it feasible for our real-world applications. To contrast these two measures of diversity, the variance among samples refers to \textit{inter-diversity}, while the variance among patches augmented from the same sample refers to \textit{intra-diversity}. We recognize that intra-diversity would inevitably suffer from redundancy in selection, as it treats each sample separately and dismisses inter-diversity among samples. An obvious solution is to inject randomness into active selection criteria, as described in Sec.~\ref{sec:randomness}. Nonetheless, a better solution is to combine inter- and intra-diversity together by computing inter-diversity locally on the smaller set of samples selected by intra-diversity. These solutions all aim at selecting sufficiently diverse samples with manageable computational complexity.

\begin{figure}[t]
\begin{center}
\includegraphics[width=1.0\linewidth]{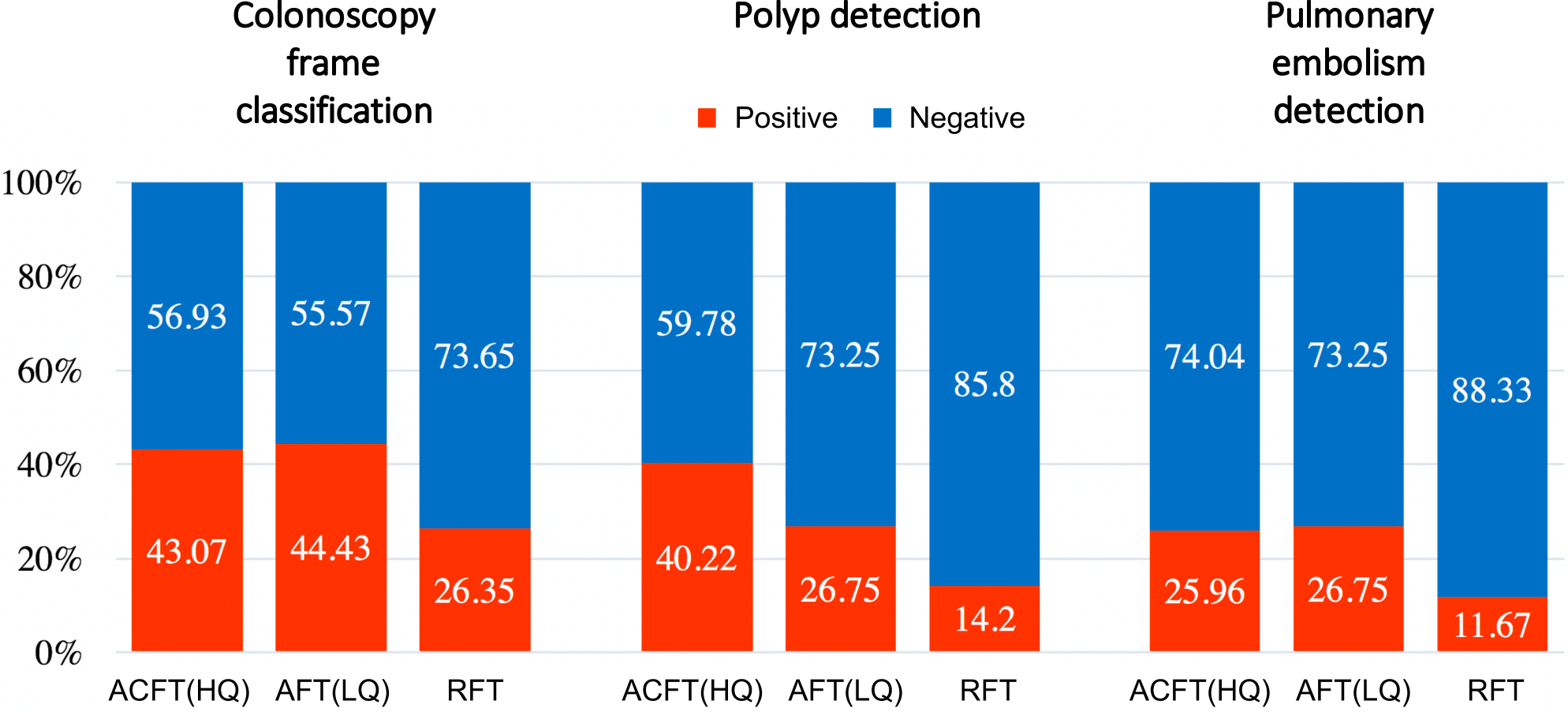}
\end{center}
\caption{The positive/negative ratio in the candidates selected by ACFT, AFT and RFT. Please note that the ratio in RFT serves as an approximation for the ratio of the entire dataset.}
\label{fig:balance_ratio}
\end{figure}

\subsection{Can actively selected samples be automatically balanced?}
\label{sec:automatically_balancing}

Data is often imbalanced in real-world applications. The images of target classes of interest, \eg certain types of diseases, only appear in a small portion of the dataset. We encounter severe imbalances in our three applications. The ratio between positives and negatives is around 1:9 in the polyp and pulmonary embolism detection. Meanwhile, the ratio is approximately 3:7 in the colonoscopy frame classification. Learning from such imbalanced datasets leads to a common issue: majority bias~\citep{aggarwal2020active}, which is a prediction bias towards majority classes over minority classes. Training data should be balanced in terms of classes~\citep{japkowicz2002class,he2009learning,buda2018systematic}. Similar to most studies in active learning literature, our proposed selection criteria are not directly designed to tackle the issue of imbalance, but they have an implicit impact on balancing the data. For instance, when the current CNN has already learned more from positive samples, the next active learning selection would be more likely to prefer those negative samples, and vice-versa. On the contrary, random selection would consistently select new samples that follow roughly the same positive/negative ratio as the entire dataset. As shown in \figurename~\ref{fig:balance_ratio}, our ACFT$_{(HQ)}$ and AFT$_{(LQ)}$ are capable of automatically balancing the selected training data. After monitoring the active selection process, ACFT$_{(HQ)}$ and AFT$_{(LQ)}$ select twice as many positives compared to random selection. This does not suggest that the number of positives and negatives must be approximately identical in the selected samples. Negative samples naturally present more contextual variance than positive ones, as negatives can contain a vast array of possibilities not including the disease of interest. It is expected that the CNN should learn more from negatives to shape the decision boundary of positives. An ideal selection should cover a sufficient variety of negatives while striking an emphasis on the positives. We believe that this accounts for the quick achievement of superior performance in imbalanced data for our ACFT$_{(HQ)}$ and AFT$_{(LQ)}$.

\begin{figure}[t]
\begin{center}
\includegraphics[width=1.0\linewidth]{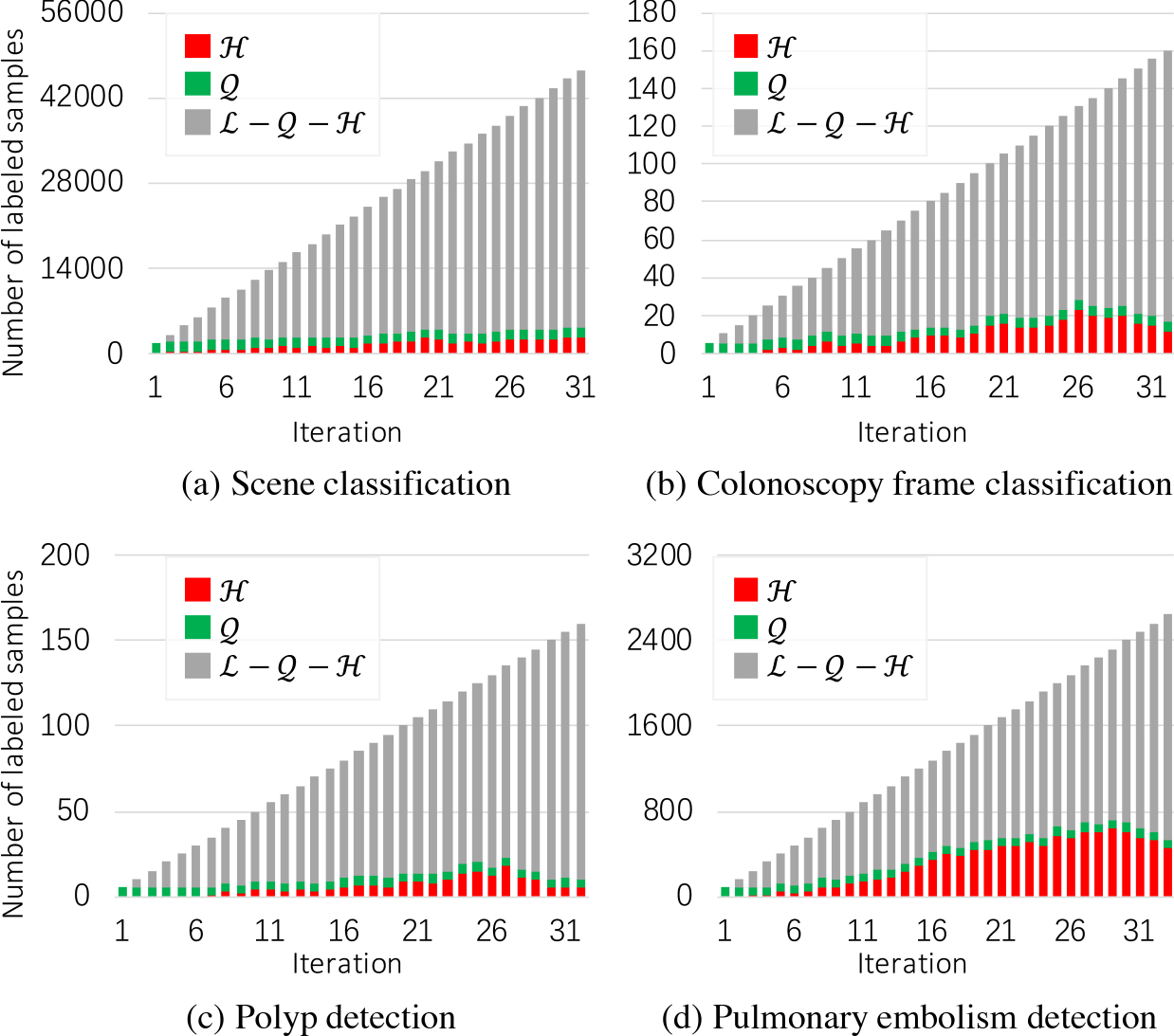}
\end{center}
\caption{
Labels are reused differently in four active learning strategies, as summarized in \tableautorefname~\ref{tab:terminology}. Specifically, the labels can be non-reused, partially reused, or 100\% reused. We plot the number of candidates along with each active learning step, including labeled candidates ($\mathcal{L}$), newly annotated candidates ($\mathcal{Q}$), and misclassified candidates ($\mathcal{H}$). As seen, by only continual fine-tuning on the hybrid data of $\mathcal{H}\bigcup\mathcal{Q}$, our ACFT significantly reduces training time through faster convergence than repeatedly fine-tuning on the entire labeled data of $\mathcal{L}\bigcup\mathcal{Q}$. Most importantly, as evidence by \tableautorefname~\ref{tab:main_results}, partially reusing labels can achieve compelling performance because it boosts performance by eliminating labeled easy candidates, focusing on hard ones, and preventing catastrophic forgetting.
}
\label{fig:label_reuse}
\end{figure}

\subsection{How to prevent model forgetting in continual learning?}

When a CNN learns from a stream of tasks continually, the learning of the new task can degrade the CNN's performance for earlier tasks~\citep{kirkpatrick2017overcoming,chen2018lifelong,parisi2019continual}. This phenomenon is called catastrophic forgetting, which was first recognized by~\citet{mccloskey1989catastrophic}. In this paper, we have also observed similar behavior in active continual fine-tuning when the CNN encounters newly selected samples. This problem might not arise if the CNN is repeatedly trained on the entire labeled set at every active learning step. 
But fully reusing the labeled samples takes a lot of resources; further especially when the labeled set gets larger and larger, the impact of the newly selected samples on the model training becomes smaller and smaller (relative to the whole labeled set).
To make the training more efficient and maximize the contribution of new data, we attempted to fine-tune the CNN only on the newly selected samples, developing the learning strategy called ACFT$_{(Q)}$. However, as seen in~\tableautorefname~\ref{tab:main_results}, ACFT$_{(Q)}$ results in a substantially unstable performance because of the catastrophic forgetting. To track the forgotten samples, we have plotted a histogram of the misclassified candidates ($\mathcal{H}$) by the current CNN against labeled candidates ($\mathcal{L}$) and newly selected candidates ($\mathcal{Q}$) in~\figurename~\ref{fig:label_reuse}. We found that if the CNN is only fine-tuned on the newly selected samples at each step, it tends to forget the samples that have been learned from previous steps. This is because new data will likely override the weights that have been learned in the past, and thus overfitting the CNN on this data and degrading the model's generalizability. Therefore, we propose to combine the newly selected ($\mathcal{Q}$) and misclassified ($\mathcal{H}$) candidates together to continual fine-tune the current CNN, which not only spotlights the power of new data to achieve the comparable performance (see~\tableautorefname~\ref{tab:main_results}: ACFT$_{(HQ)}$ vs. AFT$_{(LQ)}$), but also eases the computational cost by eliminating re-training on easy samples, focusing on hard ones, and preventing catastrophic forgetting.

{
\subsection{Is ACFT generalizable to other models?}
}
\label{sec:generalizability_architectures}

We based our experiments on AlexNet and GoogLeNet. Alternatively, deeper architectures, such as VGG~\citep{simonyan2014very}, ResNet~\citep{he2016deep}, DenseNet~\citep{huang2017densely}, and FixEfficientNet~\citep{touvron2020fixing}, could have been used and they are known to show relatively higher performance for challenging computer vision tasks.
However, the purpose of this work is not to achieve the highest performance for different medical image tasks but to answer a critical question: {\em How can annotation costs be significantly reduced when applying CNNs to medical imaging?} For this purpose, we have experimented with our three applications, demonstrating consistent patterns between AlexNet and GoogLeNet as shown in \figurename~\ref{fig:selection_approaches_comparison}. As a result, given this generalizability, we can focus on comparing the prediction patterns and learning strategies rather than running experiments on different CNN architectures. Moreover, our active selection criteria only rely on data augmentation and model prediction, without being tied to specific types of predictors. This suggests that not only various CNN architectures, but also other predictive methods---spanning old fashions (\eg SVM, Random Forests, and AdaBoost) to recent trends such as CapsuleNet~\citep{sabour2017dynamic} and Transformer~\citep{dosovitskiy2020image}---can benefit from the progress in active learning.

\subsection{Can we do better on the cold start problem?}
\label{sec:cold_start_problem}

It is crucial to intelligently select initial samples for an active learning procedure, especially for algorithms like our ACFT, which starts from a completely empty labeled dataset. Our results in~\figurename~\ref{fig:selection_approaches_comparison} and several other studies~\citep{borisov2010active,zhou2017fine,yuan2020cold} reveal that uniformly, randomly selecting initial samples from the unlabeled set could outperform active selection at the beginning. This is one of the most challenging problems in active learning, known as the \textit{cold start} problem, which is ascribed to (1) data scarcity and (2) model instability at early stages. 
First, the data distribution in randomly selected samples better reflects the original distribution of the entire dataset than in actively selected samples. Maintaining a similar distribution between training and test data is beneficial when using scarce data. The most common practice is to admit the power of randomness at the beginning and randomly select initial samples from the unlabeled set~\citep{ren2020survey}. Our paper  addresses the cold start problem by incorporating a random sampling probability with respect to the active selection criteria (as detailed in Sec.~\ref{sec:randomness}). The devised ACFT (+randomness vs. -randomness in~\figurename~\ref{fig:selection_approaches_comparison}) shows superior performance, even in early stages, with continued performance improving during the subsequent steps. 
Second, in the beginning, the CNN understandably fails to amply predict new samples, as it is trained with an inadequate number of samples. With horrible predictions, no matter how marvelous the selection criterion is, the selected samples would be unsatisfactory---as said ``garbage in garbage out''. To express meaningful CNN predictions, our paper suggests the use of pre-trained CNNs (as illustrated in Alg.~\ref{alg:ACFT}), not only initializing the CNN at the first step, but also providing fairly reasonable predictions for initial active selection. \figurename~\ref{fig:overall_result} presents encouraging results of active selection using pre-trained CNNs compared with random sampling from the unlabeled set (ACFT vs. RFT).
However, a CNN pre-trained on \textsc{ImageNet} may give poor predictions in the medical imaging domain, as it was trained from only {\em natural} images; it is associated with a large domain gap for medical images. As a result, the CNN predictions may be inconclusive or even opposite, yielding poor selection scores. Naturally, one may consider utilizing pre-trained models in the same domains to reduce this domain gap~\citep{zhou2021models,haghighi2020learning,feng2020parts2whole}. \citet{yuan2020cold} has demonstrated this idea in natural language processing by applying self-supervised language modeling to select initial samples. In the case of medical imaging, we naturally expect that self-supervised methods can also mitigate the pronounced domain gap between natural and medical imaging, offering a great starting point for selecting samples using domain-relevant image representation. More importantly, the learning objectives in self-supervised methods are applicable for discovering the most representative initial samples. 
For instance, our diversity criterion shares a similar spirit with the learning objective of BYOL~\citep{grill2020bootstrap} and of Parts2Whole~\citep{feng2020parts2whole}, as they all aim to pull together the patches augmented from the same sample.
Therefore, their objective functions could serve as an off-the-shelf measure for the power of a sample in elevating the pre-trained CNN's performance. The underlying hypothesis is that the worthiness of labeling a sample correlates with the learning objective of self-supervised pre-training. Specifically, a sample is potentially more worthy to train the CNN if it requires considerably more effort to perform the task of in-painting~\citep{pathak2016context}, restoration~\citep{zhou2021models}, contrastive learning~\citep{chen2020simple}, or colorization~\citep{zhang2016colorful}. We anticipate that self-supervised methods have great potential to accommodate the selection of initial samples by leveraging unlabeled data in the same domain, therefore, more effectively addressing the cold start problem in active learning.

\subsection{Is our consistency observation useful for other purposes?}

One of our key observations is that all patches augmented from the same sample share the same label, and thus are expected to have similar predictions by the CNN. This inherent invariance allows us to devise the diversity metric for estimating the worthiness of labeling the sample. From a broader view, the use of data consistency before and after a mixture of augmentation has played an important role in many other circumstances. In semi-supervised learning, the consistency loss serves as a bridge between labeled and unlabeled data. While the CNN is trained on labeled data, the consistency loss constrains predictions to be invariant to unlabeled data augmented in varying ways~\citep{yu2019uncertainty,cui2019semi,bortsova2019semi,fotedar2020extreme}. In self-supervised learning, the concept of consistency allows CNNs to learn transformation invariance features by either always restoring the original image from the transformed one~\citep{zhu2020rubik,zhou2021models} or explicitly pulling all patches augmented from the same image together in the feature space~\citep{feng2020parts2whole,chen2020simple,he2020momentum}. Albeit the great promises of consistency loss, automatic data augmentation inevitably generates ``noisy'' samples, jeopardizing the data consistency presumption.As an example, when an image contains objects A and B, random cropping may miss either one of the objects fully or partially, causing label inconsistency or representation inconsistency~\citep{purushwalkam2020demystifying,hinton2021represent}. Therefore, the choice of data augmentation is critical in employing the data consistency presumption. Other than data consistency, the prediction consistency of model ensembles can also calculate the diversity. For instance, \citet{gal2016dropout,gal2017deep,tsymbalov2018dropout} have proposed to estimate the prediction diversity presented in the CNN via Monte-Carlo dropout in the inference;~\citet{beluch2018power,yang2017suggestive,kuo2018cost,li2020transformation,venturini2020uncertainty} measure the prediction consistency by feeding images to multiple independent CNNs that have been trained for the same data and purpose. Unlike the data consistency in our work, their presumption is the model consistency, wherein the CNN predictions ought to be consistent if the same sample goes through the model ensembles; otherwise, this sample is considered worthy of labeling.

\section{Conclusion}
\label{sec:conclusion}

We have developed a novel method for dramatically reducing annotation cost by integrating active learning and transfer learning. Compared with the state-of-the-art random selection method~\citep{tajbakhsh2016convolutional}, our method can reduce the annotation cost by at least half for three medical applications and by more than 33$\%$ for natural image dataset \textsc{Places-3}. The superior performance of our method is attributable to eight distinct advantages, detailed in Sec.~\ref{sec:introduction}. We believe that labeling at the candidate level offers a sensible balance for our three applications, whereas labeling at the patient level would certainly enhance annotation cost reduction, but introduces more severe label noise. Labeling at the patch level compensates for additional label noise but would impose significant burdens on experts for annotation creation. 

\section*{Acknowledgments}
\label{acknowledgments}
This research has been supported partially by ASU and Mayo Clinic through a Seed Grant and an Innovation Grant, and partially by the NIH under Award Number R01HL128785. The content is solely the responsibility of the authors and does not necessarily represent the official views of the NIH. We thank S.~Tatapudi and A.~Pluhar for helping improve the writing of this paper.
The content of this paper is covered by patents pending. 

\bibliographystyle{model2-names.bst}\biboptions{authoryear}
\bibliography{refs}

\begin{thebibliography}{101}
\expandafter\ifx\csname natexlab\endcsname\relax\def\natexlab#1{#1}\fi
\providecommand{\url}[1]{\texttt{#1}}
\providecommand{\href}[2]{#2}
\providecommand{\path}[1]{#1}
\providecommand{\DOIprefix}{doi:}
\providecommand{\ArXivprefix}{arXiv:}
\providecommand{\URLprefix}{URL: }
\providecommand{\Pubmedprefix}{pmid:}
\providecommand{\doi}[1]{\href{http://dx.doi.org/#1}{\path{#1}}}
\providecommand{\Pubmed}[1]{\href{pmid:#1}{\path{#1}}}
\providecommand{\bibinfo}[2]{#2}
\ifx\xfnm\relax \def\xfnm[#1]{\unskip,\space#1}\fi
\bibitem[{Aggarwal et~al.(2020)Aggarwal, Popescu and
  Hudelot}]{aggarwal2020active}
\bibinfo{author}{Aggarwal, U.}, \bibinfo{author}{Popescu, A.},
  \bibinfo{author}{Hudelot, C.}, \bibinfo{year}{2020}.
\newblock \bibinfo{title}{Active learning for imbalanced datasets}, in:
  \bibinfo{booktitle}{Proceedings of the IEEE/CVF Winter Conference on
  Applications of Computer Vision}, pp. \bibinfo{pages}{1428--1437}.
\bibitem[{Ardila et~al.(2019)Ardila, Kiraly, Bharadwaj, Choi, Reicher, Peng,
  Tse, Etemadi, Ye, Corrado et~al.}]{ardila2019end}
\bibinfo{author}{Ardila, D.}, \bibinfo{author}{Kiraly, A.P.},
  \bibinfo{author}{Bharadwaj, S.}, \bibinfo{author}{Choi, B.},
  \bibinfo{author}{Reicher, J.J.}, \bibinfo{author}{Peng, L.},
  \bibinfo{author}{Tse, D.}, \bibinfo{author}{Etemadi, M.},
  \bibinfo{author}{Ye, W.}, \bibinfo{author}{Corrado, G.}, et~al.,
  \bibinfo{year}{2019}.
\newblock \bibinfo{title}{End-to-end lung cancer screening with
  three-dimensional deep learning on low-dose chest computed tomography}.
\newblock \bibinfo{journal}{Nature medicine} \bibinfo{volume}{25},
  \bibinfo{pages}{954--961}.
\bibitem[{Azizi et~al.(2021)Azizi, Mustafa, Ryan, Beaver, Freyberg, Deaton,
  Loh, Karthikesalingam, Kornblith, Chen et~al.}]{azizi2021big}
\bibinfo{author}{Azizi, S.}, \bibinfo{author}{Mustafa, B.},
  \bibinfo{author}{Ryan, F.}, \bibinfo{author}{Beaver, Z.},
  \bibinfo{author}{Freyberg, J.}, \bibinfo{author}{Deaton, J.},
  \bibinfo{author}{Loh, A.}, \bibinfo{author}{Karthikesalingam, A.},
  \bibinfo{author}{Kornblith, S.}, \bibinfo{author}{Chen, T.}, et~al.,
  \bibinfo{year}{2021}.
\newblock \bibinfo{title}{Big self-supervised models advance medical image
  classification}.
\newblock \bibinfo{journal}{arXiv preprint arXiv:2101.05224} .
\bibitem[{Balcan et~al.(2007)Balcan, Broder and Zhang}]{balcan2007margin}
\bibinfo{author}{Balcan, M.F.}, \bibinfo{author}{Broder, A.},
  \bibinfo{author}{Zhang, T.}, \bibinfo{year}{2007}.
\newblock \bibinfo{title}{Margin based active learning}, in:
  \bibinfo{booktitle}{International Conference on Computational Learning
  Theory}, \bibinfo{organization}{Springer}. pp. \bibinfo{pages}{35--50}.
\bibitem[{Beluch et~al.(2018)Beluch, Genewein, N{\"u}rnberger and
  K{\"o}hler}]{beluch2018power}
\bibinfo{author}{Beluch, W.H.}, \bibinfo{author}{Genewein, T.},
  \bibinfo{author}{N{\"u}rnberger, A.}, \bibinfo{author}{K{\"o}hler, J.M.},
  \bibinfo{year}{2018}.
\newblock \bibinfo{title}{The power of ensembles for active learning in image
  classification}, in: \bibinfo{booktitle}{Proceedings of the IEEE Conference
  on Computer Vision and Pattern Recognition}, pp. \bibinfo{pages}{9368--9377}.
\bibitem[{Borisov et~al.(2010)Borisov, Tuv and Runger}]{borisov2010active}
\bibinfo{author}{Borisov, A.}, \bibinfo{author}{Tuv, E.},
  \bibinfo{author}{Runger, G.}, \bibinfo{year}{2010}.
\newblock \bibinfo{title}{Active batch learning with stochastic query by
  forest}, in: \bibinfo{booktitle}{JMLR: Workshop and Conference Proceedings
  (2010)}, \bibinfo{organization}{Citeseer}.
\bibitem[{Bortsova et~al.(2019)Bortsova, Dubost, Hogeweg, Katramados and
  de~Bruijne}]{bortsova2019semi}
\bibinfo{author}{Bortsova, G.}, \bibinfo{author}{Dubost, F.},
  \bibinfo{author}{Hogeweg, L.}, \bibinfo{author}{Katramados, I.},
  \bibinfo{author}{de~Bruijne, M.}, \bibinfo{year}{2019}.
\newblock \bibinfo{title}{Semi-supervised medical image segmentation via
  learning consistency under transformations}, in:
  \bibinfo{booktitle}{International Conference on Medical Image Computing and
  Computer-Assisted Intervention}, \bibinfo{organization}{Springer}. pp.
  \bibinfo{pages}{810--818}.
\bibitem[{Buda et~al.(2018)Buda, Maki and Mazurowski}]{buda2018systematic}
\bibinfo{author}{Buda, M.}, \bibinfo{author}{Maki, A.},
  \bibinfo{author}{Mazurowski, M.A.}, \bibinfo{year}{2018}.
\newblock \bibinfo{title}{A systematic study of the class imbalance problem in
  convolutional neural networks}.
\newblock \bibinfo{journal}{Neural Networks} \bibinfo{volume}{106},
  \bibinfo{pages}{249--259}.
\bibitem[{Caron et~al.(2020)Caron, Misra, Mairal, Goyal, Bojanowski and
  Joulin}]{caron2020unsupervised}
\bibinfo{author}{Caron, M.}, \bibinfo{author}{Misra, I.},
  \bibinfo{author}{Mairal, J.}, \bibinfo{author}{Goyal, P.},
  \bibinfo{author}{Bojanowski, P.}, \bibinfo{author}{Joulin, A.},
  \bibinfo{year}{2020}.
\newblock \bibinfo{title}{Unsupervised learning of visual features by
  contrasting cluster assignments}.
\newblock \bibinfo{journal}{arXiv preprint arXiv:2006.09882} .
\bibitem[{Chakraborty et~al.(2015)Chakraborty, Balasubramanian, Sun,
  Panchanathan and Ye}]{chakraborty2015active}
\bibinfo{author}{Chakraborty, S.}, \bibinfo{author}{Balasubramanian, V.},
  \bibinfo{author}{Sun, Q.}, \bibinfo{author}{Panchanathan, S.},
  \bibinfo{author}{Ye, J.}, \bibinfo{year}{2015}.
\newblock \bibinfo{title}{Active batch selection via convex relaxations with
  guaranteed solution bounds}.
\newblock \bibinfo{journal}{IEEE transactions on pattern analysis and machine
  intelligence} \bibinfo{volume}{37}, \bibinfo{pages}{1945--1958}.
\bibitem[{Chen et~al.(2019)Chen, Ma and Zheng}]{chen2019med3d}
\bibinfo{author}{Chen, S.}, \bibinfo{author}{Ma, K.}, \bibinfo{author}{Zheng,
  Y.}, \bibinfo{year}{2019}.
\newblock \bibinfo{title}{Med3d: Transfer learning for 3d medical image
  analysis}.
\newblock \bibinfo{journal}{arXiv preprint arXiv:1904.00625} .
\bibitem[{Chen et~al.(2020)Chen, Kornblith, Norouzi and
  Hinton}]{chen2020simple}
\bibinfo{author}{Chen, T.}, \bibinfo{author}{Kornblith, S.},
  \bibinfo{author}{Norouzi, M.}, \bibinfo{author}{Hinton, G.},
  \bibinfo{year}{2020}.
\newblock \bibinfo{title}{A simple framework for contrastive learning of visual
  representations}.
\newblock \bibinfo{journal}{arXiv preprint arXiv:2002.05709} .
\bibitem[{Chen and He(2020)}]{chen2020exploring}
\bibinfo{author}{Chen, X.}, \bibinfo{author}{He, K.}, \bibinfo{year}{2020}.
\newblock \bibinfo{title}{Exploring simple siamese representation learning}.
\newblock \bibinfo{journal}{arXiv preprint arXiv:2011.10566} .
\bibitem[{Chen and Liu(2018)}]{chen2018lifelong}
\bibinfo{author}{Chen, Z.}, \bibinfo{author}{Liu, B.}, \bibinfo{year}{2018}.
\newblock \bibinfo{title}{Lifelong machine learning}.
\newblock \bibinfo{journal}{Synthesis Lectures on Artificial Intelligence and
  Machine Learning} \bibinfo{volume}{12}, \bibinfo{pages}{1--207}.
\bibitem[{Cui et~al.(2019)Cui, Liu, Li, Guo, Li, Li, Wang, Zeng and
  Ye}]{cui2019semi}
\bibinfo{author}{Cui, W.}, \bibinfo{author}{Liu, Y.}, \bibinfo{author}{Li, Y.},
  \bibinfo{author}{Guo, M.}, \bibinfo{author}{Li, Y.}, \bibinfo{author}{Li,
  X.}, \bibinfo{author}{Wang, T.}, \bibinfo{author}{Zeng, X.},
  \bibinfo{author}{Ye, C.}, \bibinfo{year}{2019}.
\newblock \bibinfo{title}{Semi-supervised brain lesion segmentation with an
  adapted mean teacher model}, in: \bibinfo{booktitle}{International Conference
  on Information Processing in Medical Imaging},
  \bibinfo{organization}{Springer}. pp. \bibinfo{pages}{554--565}.
\bibitem[{Culotta and McCallum(2005)}]{culotta2005reducing}
\bibinfo{author}{Culotta, A.}, \bibinfo{author}{McCallum, A.},
  \bibinfo{year}{2005}.
\newblock \bibinfo{title}{Reducing labeling effort for structured prediction
  tasks}, in: \bibinfo{booktitle}{AAAI}, pp. \bibinfo{pages}{746--751}.
\bibitem[{Dagan and Engelson(1995)}]{dagan1995committee}
\bibinfo{author}{Dagan, I.}, \bibinfo{author}{Engelson, S.P.},
  \bibinfo{year}{1995}.
\newblock \bibinfo{title}{Committee-based sampling for training probabilistic
  classifiers}, in: \bibinfo{booktitle}{Machine Learning Proceedings 1995}.
  \bibinfo{publisher}{Elsevier}, pp. \bibinfo{pages}{150--157}.
\bibitem[{Deng et~al.(2009)Deng, Dong, Socher, Li, Li and
  Fei-Fei}]{deng2009imagenet}
\bibinfo{author}{Deng, J.}, \bibinfo{author}{Dong, W.},
  \bibinfo{author}{Socher, R.}, \bibinfo{author}{Li, L.J.},
  \bibinfo{author}{Li, K.}, \bibinfo{author}{Fei-Fei, L.},
  \bibinfo{year}{2009}.
\newblock \bibinfo{title}{Imagenet: A large-scale hierarchical image database},
  in: \bibinfo{booktitle}{Proceedings of the IEEE Conference on Computer Vision
  and Pattern Recognition}, \bibinfo{organization}{IEEE}. pp.
  \bibinfo{pages}{248--255}.
\bibitem[{Ding et~al.(2018)Ding, Sohn, Kawczynski, Trivedi, Harnish, Jenkins,
  Lituiev, Copeland, Aboian, Mari~Aparici et~al.}]{ding2018deep}
\bibinfo{author}{Ding, Y.}, \bibinfo{author}{Sohn, J.H.},
  \bibinfo{author}{Kawczynski, M.G.}, \bibinfo{author}{Trivedi, H.},
  \bibinfo{author}{Harnish, R.}, \bibinfo{author}{Jenkins, N.W.},
  \bibinfo{author}{Lituiev, D.}, \bibinfo{author}{Copeland, T.P.},
  \bibinfo{author}{Aboian, M.S.}, \bibinfo{author}{Mari~Aparici, C.}, et~al.,
  \bibinfo{year}{2018}.
\newblock \bibinfo{title}{A deep learning model to predict a diagnosis of
  alzheimer disease by using 18f-fdg pet of the brain}.
\newblock \bibinfo{journal}{Radiology} \bibinfo{volume}{290},
  \bibinfo{pages}{456--464}.
\bibitem[{Dosovitskiy et~al.(2020)Dosovitskiy, Beyer, Kolesnikov, Weissenborn,
  Zhai, Unterthiner, Dehghani, Minderer, Heigold, Gelly
  et~al.}]{dosovitskiy2020image}
\bibinfo{author}{Dosovitskiy, A.}, \bibinfo{author}{Beyer, L.},
  \bibinfo{author}{Kolesnikov, A.}, \bibinfo{author}{Weissenborn, D.},
  \bibinfo{author}{Zhai, X.}, \bibinfo{author}{Unterthiner, T.},
  \bibinfo{author}{Dehghani, M.}, \bibinfo{author}{Minderer, M.},
  \bibinfo{author}{Heigold, G.}, \bibinfo{author}{Gelly, S.}, et~al.,
  \bibinfo{year}{2020}.
\newblock \bibinfo{title}{An image is worth 16x16 words: Transformers for image
  recognition at scale}.
\newblock \bibinfo{journal}{arXiv preprint arXiv:2010.11929} .
\bibitem[{Esteva et~al.(2017)Esteva, Kuprel, Novoa, Ko, Swetter, Blau and
  Thrun}]{esteva2017dermatologist}
\bibinfo{author}{Esteva, A.}, \bibinfo{author}{Kuprel, B.},
  \bibinfo{author}{Novoa, R.A.}, \bibinfo{author}{Ko, J.},
  \bibinfo{author}{Swetter, S.M.}, \bibinfo{author}{Blau, H.M.},
  \bibinfo{author}{Thrun, S.}, \bibinfo{year}{2017}.
\newblock \bibinfo{title}{Dermatologist-level classification of skin cancer
  with deep neural networks}.
\newblock \bibinfo{journal}{Nature} \bibinfo{volume}{542},
  \bibinfo{pages}{115}.
\bibitem[{Esteva et~al.(2019)Esteva, Robicquet, Ramsundar, Kuleshov, DePristo,
  Chou, Cui, Corrado, Thrun and Dean}]{esteva2019guide}
\bibinfo{author}{Esteva, A.}, \bibinfo{author}{Robicquet, A.},
  \bibinfo{author}{Ramsundar, B.}, \bibinfo{author}{Kuleshov, V.},
  \bibinfo{author}{DePristo, M.}, \bibinfo{author}{Chou, K.},
  \bibinfo{author}{Cui, C.}, \bibinfo{author}{Corrado, G.},
  \bibinfo{author}{Thrun, S.}, \bibinfo{author}{Dean, J.},
  \bibinfo{year}{2019}.
\newblock \bibinfo{title}{A guide to deep learning in healthcare}.
\newblock \bibinfo{journal}{Nature medicine} \bibinfo{volume}{25},
  \bibinfo{pages}{24--29}.
\bibitem[{Feng et~al.(2020)Feng, Zhou, Gotway and Liang}]{feng2020parts2whole}
\bibinfo{author}{Feng, R.}, \bibinfo{author}{Zhou, Z.},
  \bibinfo{author}{Gotway, M.B.}, \bibinfo{author}{Liang, J.},
  \bibinfo{year}{2020}.
\newblock \bibinfo{title}{Parts2whole: Self-supervised contrastive learning via
  reconstruction}, in: \bibinfo{booktitle}{Domain Adaptation and Representation
  Transfer, and Distributed and Collaborative Learning}.
  \bibinfo{publisher}{Springer}, pp. \bibinfo{pages}{85--95}.
\bibitem[{Fotedar et~al.(2020)Fotedar, Tajbakhsh, Ananth and
  Ding}]{fotedar2020extreme}
\bibinfo{author}{Fotedar, G.}, \bibinfo{author}{Tajbakhsh, N.},
  \bibinfo{author}{Ananth, S.}, \bibinfo{author}{Ding, X.},
  \bibinfo{year}{2020}.
\newblock \bibinfo{title}{Extreme consistency: Overcoming annotation scarcity
  and domain shifts}, in: \bibinfo{booktitle}{International Conference on
  Medical Image Computing and Computer-Assisted Intervention},
  \bibinfo{organization}{Springer}. pp. \bibinfo{pages}{699--709}.
\bibitem[{Gal and Ghahramani(2016)}]{gal2016dropout}
\bibinfo{author}{Gal, Y.}, \bibinfo{author}{Ghahramani, Z.},
  \bibinfo{year}{2016}.
\newblock \bibinfo{title}{Dropout as a bayesian approximation: Representing
  model uncertainty in deep learning}, in: \bibinfo{booktitle}{international
  conference on machine learning}, \bibinfo{organization}{PMLR}. pp.
  \bibinfo{pages}{1050--1059}.
\bibitem[{Gal et~al.(2017)Gal, Islam and Ghahramani}]{gal2017deep}
\bibinfo{author}{Gal, Y.}, \bibinfo{author}{Islam, R.},
  \bibinfo{author}{Ghahramani, Z.}, \bibinfo{year}{2017}.
\newblock \bibinfo{title}{Deep bayesian active learning with image data}, in:
  \bibinfo{booktitle}{International Conference on Machine Learning},
  \bibinfo{organization}{PMLR}. pp. \bibinfo{pages}{1183--1192}.
\bibitem[{Grill et~al.(2020)Grill, Strub, Altch{\'e}, Tallec, Richemond,
  Buchatskaya, Doersch, Pires, Guo, Azar et~al.}]{grill2020bootstrap}
\bibinfo{author}{Grill, J.B.}, \bibinfo{author}{Strub, F.},
  \bibinfo{author}{Altch{\'e}, F.}, \bibinfo{author}{Tallec, C.},
  \bibinfo{author}{Richemond, P.H.}, \bibinfo{author}{Buchatskaya, E.},
  \bibinfo{author}{Doersch, C.}, \bibinfo{author}{Pires, B.A.},
  \bibinfo{author}{Guo, Z.D.}, \bibinfo{author}{Azar, M.G.}, et~al.,
  \bibinfo{year}{2020}.
\newblock \bibinfo{title}{Bootstrap your own latent: A new approach to
  self-supervised learning}.
\newblock \bibinfo{journal}{arXiv preprint arXiv:2006.07733} .
\bibitem[{Guan and Huang(2018)}]{guan2018multi}
\bibinfo{author}{Guan, Q.}, \bibinfo{author}{Huang, Y.}, \bibinfo{year}{2018}.
\newblock \bibinfo{title}{Multi-label chest x-ray image classification via
  category-wise residual attention learning}.
\newblock \bibinfo{journal}{Pattern Recognition Letters} .
\bibitem[{Guendel et~al.(2018)Guendel, Grbic, Georgescu, Liu, Maier and
  Comaniciu}]{guendel2018learning}
\bibinfo{author}{Guendel, S.}, \bibinfo{author}{Grbic, S.},
  \bibinfo{author}{Georgescu, B.}, \bibinfo{author}{Liu, S.},
  \bibinfo{author}{Maier, A.}, \bibinfo{author}{Comaniciu, D.},
  \bibinfo{year}{2018}.
\newblock \bibinfo{title}{Learning to recognize abnormalities in chest x-rays
  with location-aware dense networks}, in: \bibinfo{booktitle}{Iberoamerican
  Congress on Pattern Recognition}, \bibinfo{organization}{Springer}. pp.
  \bibinfo{pages}{757--765}.
\bibitem[{Guyon et~al.(2011)Guyon, Cawley, Dror and Lemaire}]{guyon2011results}
\bibinfo{author}{Guyon, I.}, \bibinfo{author}{Cawley, G.C.},
  \bibinfo{author}{Dror, G.}, \bibinfo{author}{Lemaire, V.},
  \bibinfo{year}{2011}.
\newblock \bibinfo{title}{Results of the active learning challenge}, in:
  \bibinfo{booktitle}{Active Learning and Experimental Design workshop In
  conjunction with AISTATS 2010}, pp. \bibinfo{pages}{19--45}.
\bibitem[{Haghighi et~al.(2020)Haghighi, Taher, Zhou, Gotway and
  Liang}]{haghighi2020learning}
\bibinfo{author}{Haghighi, F.}, \bibinfo{author}{Taher, M.R.H.},
  \bibinfo{author}{Zhou, Z.}, \bibinfo{author}{Gotway, M.B.},
  \bibinfo{author}{Liang, J.}, \bibinfo{year}{2020}.
\newblock \bibinfo{title}{Learning semantics-enriched representation via
  self-discovery, self-classification, and self-restoration}, in:
  \bibinfo{booktitle}{International Conference on Medical Image Computing and
  Computer-Assisted Intervention}, \bibinfo{organization}{Springer}. pp.
  \bibinfo{pages}{137--147}.
\bibitem[{He and Garcia(2009)}]{he2009learning}
\bibinfo{author}{He, H.}, \bibinfo{author}{Garcia, E.A.}, \bibinfo{year}{2009}.
\newblock \bibinfo{title}{Learning from imbalanced data}.
\newblock \bibinfo{journal}{IEEE Transactions on knowledge and data
  engineering} \bibinfo{volume}{21}, \bibinfo{pages}{1263--1284}.
\bibitem[{He et~al.(2020)He, Fan, Wu, Xie and Girshick}]{he2020momentum}
\bibinfo{author}{He, K.}, \bibinfo{author}{Fan, H.}, \bibinfo{author}{Wu, Y.},
  \bibinfo{author}{Xie, S.}, \bibinfo{author}{Girshick, R.},
  \bibinfo{year}{2020}.
\newblock \bibinfo{title}{Momentum contrast for unsupervised visual
  representation learning}, in: \bibinfo{booktitle}{Proceedings of the IEEE/CVF
  Conference on Computer Vision and Pattern Recognition}, pp.
  \bibinfo{pages}{9729--9738}.
\bibitem[{He et~al.(2016)He, Zhang, Ren and Sun}]{he2016deep}
\bibinfo{author}{He, K.}, \bibinfo{author}{Zhang, X.}, \bibinfo{author}{Ren,
  S.}, \bibinfo{author}{Sun, J.}, \bibinfo{year}{2016}.
\newblock \bibinfo{title}{Deep residual learning for image recognition}, in:
  \bibinfo{booktitle}{Proceedings of the IEEE Conference on Computer Vision and
  Pattern Recognition}, pp. \bibinfo{pages}{770--778}.
\bibitem[{Hino(2020)}]{hino2020active}
\bibinfo{author}{Hino, H.}, \bibinfo{year}{2020}.
\newblock \bibinfo{title}{Active learning: Problem settings and recent
  developments}.
\newblock \bibinfo{journal}{arXiv preprint arXiv:2012.04225} .
\bibitem[{Hinton(2021)}]{hinton2021represent}
\bibinfo{author}{Hinton, G.}, \bibinfo{year}{2021}.
\newblock \bibinfo{title}{How to represent part-whole hierarchies in a neural
  network}.
\newblock \bibinfo{journal}{arXiv preprint arXiv:2102.12627} .
\bibitem[{Holub et~al.(2008)Holub, Perona and Burl}]{holub2008entropy}
\bibinfo{author}{Holub, A.}, \bibinfo{author}{Perona, P.},
  \bibinfo{author}{Burl, M.C.}, \bibinfo{year}{2008}.
\newblock \bibinfo{title}{Entropy-based active learning for object
  recognition}, in: \bibinfo{booktitle}{2008 IEEE Computer Society Conference
  on Computer Vision and Pattern Recognition Workshops},
  \bibinfo{organization}{IEEE}. pp. \bibinfo{pages}{1--8}.
\bibitem[{Huang et~al.(2017)Huang, Liu, Weinberger and van~der
  Maaten}]{huang2017densely}
\bibinfo{author}{Huang, G.}, \bibinfo{author}{Liu, Z.},
  \bibinfo{author}{Weinberger, K.Q.}, \bibinfo{author}{van~der Maaten, L.},
  \bibinfo{year}{2017}.
\newblock \bibinfo{title}{Densely connected convolutional networks}, in:
  \bibinfo{booktitle}{Proceedings of the IEEE Conference on Computer Vision and
  Pattern Recognition}, p.~\bibinfo{pages}{3}.
\bibitem[{Huang et~al.(2020)Huang, Kothari, Banerjee, Chute, Ball, Borus,
  Huang, Patel, Rajpurkar, Irvin et~al.}]{huang2020penet}
\bibinfo{author}{Huang, S.C.}, \bibinfo{author}{Kothari, T.},
  \bibinfo{author}{Banerjee, I.}, \bibinfo{author}{Chute, C.},
  \bibinfo{author}{Ball, R.L.}, \bibinfo{author}{Borus, N.},
  \bibinfo{author}{Huang, A.}, \bibinfo{author}{Patel, B.N.},
  \bibinfo{author}{Rajpurkar, P.}, \bibinfo{author}{Irvin, J.}, et~al.,
  \bibinfo{year}{2020}.
\newblock \bibinfo{title}{Penet—a scalable deep-learning model for automated
  diagnosis of pulmonary embolism using volumetric ct imaging}.
\newblock \bibinfo{journal}{npj Digital Medicine} \bibinfo{volume}{3},
  \bibinfo{pages}{1--9}.
\bibitem[{Irvin et~al.(2019)Irvin, Rajpurkar, Ko, Yu, Ciurea-Ilcus, Chute,
  Marklund, Haghgoo, Ball, Shpanskaya et~al.}]{irvin2019chexpert}
\bibinfo{author}{Irvin, J.}, \bibinfo{author}{Rajpurkar, P.},
  \bibinfo{author}{Ko, M.}, \bibinfo{author}{Yu, Y.},
  \bibinfo{author}{Ciurea-Ilcus, S.}, \bibinfo{author}{Chute, C.},
  \bibinfo{author}{Marklund, H.}, \bibinfo{author}{Haghgoo, B.},
  \bibinfo{author}{Ball, R.}, \bibinfo{author}{Shpanskaya, K.}, et~al.,
  \bibinfo{year}{2019}.
\newblock \bibinfo{title}{Chexpert: A large chest radiograph dataset with
  uncertainty labels and expert comparison}, in:
  \bibinfo{booktitle}{Proceedings of the AAAI Conference on Artificial
  Intelligence}, pp. \bibinfo{pages}{590--597}.
\bibitem[{Isensee et~al.(2021)Isensee, Jaeger, Kohl, Petersen and
  Maier-Hein}]{isensee2021nnu}
\bibinfo{author}{Isensee, F.}, \bibinfo{author}{Jaeger, P.F.},
  \bibinfo{author}{Kohl, S.A.}, \bibinfo{author}{Petersen, J.},
  \bibinfo{author}{Maier-Hein, K.H.}, \bibinfo{year}{2021}.
\newblock \bibinfo{title}{nnu-net: a self-configuring method for deep
  learning-based biomedical image segmentation}.
\newblock \bibinfo{journal}{Nature Methods} \bibinfo{volume}{18},
  \bibinfo{pages}{203--211}.
\bibitem[{Japkowicz and Stephen(2002)}]{japkowicz2002class}
\bibinfo{author}{Japkowicz, N.}, \bibinfo{author}{Stephen, S.},
  \bibinfo{year}{2002}.
\newblock \bibinfo{title}{The class imbalance problem: A systematic study}.
\newblock \bibinfo{journal}{Intelligent data analysis} \bibinfo{volume}{6},
  \bibinfo{pages}{429--449}.
\bibitem[{K{\"a}ding et~al.(2016)K{\"a}ding, Rodner, Freytag and
  Denzler}]{kading2016fine}
\bibinfo{author}{K{\"a}ding, C.}, \bibinfo{author}{Rodner, E.},
  \bibinfo{author}{Freytag, A.}, \bibinfo{author}{Denzler, J.},
  \bibinfo{year}{2016}.
\newblock \bibinfo{title}{Fine-tuning deep neural networks in continuous
  learning scenarios}, in: \bibinfo{booktitle}{Asian Conference on Computer
  Vision}, \bibinfo{organization}{Springer}. pp. \bibinfo{pages}{588--605}.
\bibitem[{Kirkpatrick et~al.(2017)Kirkpatrick, Pascanu, Rabinowitz, Veness,
  Desjardins, Rusu, Milan, Quan, Ramalho, Grabska-Barwinska
  et~al.}]{kirkpatrick2017overcoming}
\bibinfo{author}{Kirkpatrick, J.}, \bibinfo{author}{Pascanu, R.},
  \bibinfo{author}{Rabinowitz, N.}, \bibinfo{author}{Veness, J.},
  \bibinfo{author}{Desjardins, G.}, \bibinfo{author}{Rusu, A.A.},
  \bibinfo{author}{Milan, K.}, \bibinfo{author}{Quan, J.},
  \bibinfo{author}{Ramalho, T.}, \bibinfo{author}{Grabska-Barwinska, A.},
  et~al., \bibinfo{year}{2017}.
\newblock \bibinfo{title}{Overcoming catastrophic forgetting in neural
  networks}.
\newblock \bibinfo{journal}{Proceedings of the national academy of sciences}
  \bibinfo{volume}{114}, \bibinfo{pages}{3521--3526}.
\bibitem[{Krizhevsky et~al.(2012)Krizhevsky, Sutskever and
  Hinton}]{krizhevsky2012imagenet}
\bibinfo{author}{Krizhevsky, A.}, \bibinfo{author}{Sutskever, I.},
  \bibinfo{author}{Hinton, G.E.}, \bibinfo{year}{2012}.
\newblock \bibinfo{title}{Imagenet classification with deep convolutional
  neural networks}, in: \bibinfo{booktitle}{Advances in neural information
  processing systems}, pp. \bibinfo{pages}{1097--1105}.
\bibitem[{Kukar(2003)}]{kukar2003transductive}
\bibinfo{author}{Kukar, M.}, \bibinfo{year}{2003}.
\newblock \bibinfo{title}{Transductive reliability estimation for medical
  diagnosis}.
\newblock \bibinfo{journal}{Artificial Intelligence in Medicine}
  \bibinfo{volume}{29}, \bibinfo{pages}{81--106}.
\bibitem[{Kulick et~al.(2014)Kulick, Lieck, Toussaint
  et~al.}]{kulick2014active}
\bibinfo{author}{Kulick, J.}, \bibinfo{author}{Lieck, R.},
  \bibinfo{author}{Toussaint, M.}, et~al., \bibinfo{year}{2014}.
\newblock \bibinfo{title}{Active learning of hyperparameters: An expected cross
  entropy criterion for active model selection}.
\newblock \bibinfo{journal}{ArXiv e-prints} .
\bibitem[{Kuo et~al.(2018)Kuo, H{\"a}ne, Yuh, Mukherjee and
  Malik}]{kuo2018cost}
\bibinfo{author}{Kuo, W.}, \bibinfo{author}{H{\"a}ne, C.},
  \bibinfo{author}{Yuh, E.}, \bibinfo{author}{Mukherjee, P.},
  \bibinfo{author}{Malik, J.}, \bibinfo{year}{2018}.
\newblock \bibinfo{title}{Cost-sensitive active learning for intracranial
  hemorrhage detection}, in: \bibinfo{booktitle}{International Conference on
  Medical Image Computing and Computer-Assisted Intervention},
  \bibinfo{organization}{Springer}. pp. \bibinfo{pages}{715--723}.
\bibitem[{LeCun et~al.(2015)LeCun, Bengio and Hinton}]{lecun2015deep}
\bibinfo{author}{LeCun, Y.}, \bibinfo{author}{Bengio, Y.},
  \bibinfo{author}{Hinton, G.}, \bibinfo{year}{2015}.
\newblock \bibinfo{title}{Deep learning}.
\newblock \bibinfo{journal}{nature} \bibinfo{volume}{521},
  \bibinfo{pages}{436}.
\bibitem[{Li and Guo(2013)}]{li2013adaptive}
\bibinfo{author}{Li, X.}, \bibinfo{author}{Guo, Y.}, \bibinfo{year}{2013}.
\newblock \bibinfo{title}{Adaptive active learning for image classification},
  in: \bibinfo{booktitle}{Proceedings of the IEEE Conference on Computer Vision
  and Pattern Recognition}, pp. \bibinfo{pages}{859--866}.
\bibitem[{Li et~al.(2020)Li, Yu, Chen, Fu, Xing and
  Heng}]{li2020transformation}
\bibinfo{author}{Li, X.}, \bibinfo{author}{Yu, L.}, \bibinfo{author}{Chen, H.},
  \bibinfo{author}{Fu, C.W.}, \bibinfo{author}{Xing, L.},
  \bibinfo{author}{Heng, P.A.}, \bibinfo{year}{2020}.
\newblock \bibinfo{title}{Transformation-consistent self-ensembling model for
  semisupervised medical image segmentation}.
\newblock \bibinfo{journal}{IEEE Transactions on Neural Networks and Learning
  Systems} .
\bibitem[{Lu et~al.(2017)Lu, Zheng, Carneiro and Yang}]{lu2017deep}
\bibinfo{author}{Lu, L.}, \bibinfo{author}{Zheng, Y.},
  \bibinfo{author}{Carneiro, G.}, \bibinfo{author}{Yang, L.},
  \bibinfo{year}{2017}.
\newblock \bibinfo{title}{Deep learning and convolutional neural networks for
  medical image computing}.
\newblock \bibinfo{journal}{Advances in Computer Vision and Pattern
  Recognition} .
\bibitem[{Ma et~al.(2019)Ma, Zhou, Chen, Lu and Zhao}]{ma2019multi}
\bibinfo{author}{Ma, Y.}, \bibinfo{author}{Zhou, Q.}, \bibinfo{author}{Chen,
  X.}, \bibinfo{author}{Lu, H.}, \bibinfo{author}{Zhao, Y.},
  \bibinfo{year}{2019}.
\newblock \bibinfo{title}{Multi-attention network for thoracic disease
  classification and localization}, in: \bibinfo{booktitle}{ICASSP 2019-2019
  IEEE International Conference on Acoustics, Speech and Signal Processing
  (ICASSP)}, \bibinfo{organization}{IEEE}. pp. \bibinfo{pages}{1378--1382}.
\bibitem[{Mahapatra et~al.(2018)Mahapatra, Bozorgtabar, Thiran and
  Reyes}]{mahapatra2018efficient}
\bibinfo{author}{Mahapatra, D.}, \bibinfo{author}{Bozorgtabar, B.},
  \bibinfo{author}{Thiran, J.P.}, \bibinfo{author}{Reyes, M.},
  \bibinfo{year}{2018}.
\newblock \bibinfo{title}{Efficient active learning for image classification
  and segmentation using a sample selection and conditional generative
  adversarial network}, in: \bibinfo{booktitle}{International Conference on
  Medical Image Computing and Computer-Assisted Intervention},
  \bibinfo{organization}{Springer}. pp. \bibinfo{pages}{580--588}.
\bibitem[{McCallumzy and Nigamy(1998)}]{mccallumzy1998employing}
\bibinfo{author}{McCallumzy, A.K.}, \bibinfo{author}{Nigamy, K.},
  \bibinfo{year}{1998}.
\newblock \bibinfo{title}{Employing em and pool-based active learning for text
  classification}, in: \bibinfo{booktitle}{Proc. International Conference on
  Machine Learning (ICML)}, \bibinfo{organization}{Citeseer}. pp.
  \bibinfo{pages}{359--367}.
\bibitem[{McCloskey and Cohen(1989)}]{mccloskey1989catastrophic}
\bibinfo{author}{McCloskey, M.}, \bibinfo{author}{Cohen, N.J.},
  \bibinfo{year}{1989}.
\newblock \bibinfo{title}{Catastrophic interference in connectionist networks:
  The sequential learning problem}, in: \bibinfo{booktitle}{Psychology of
  learning and motivation}. \bibinfo{publisher}{Elsevier}.
  volume~\bibinfo{volume}{24}, pp. \bibinfo{pages}{109--165}.
\bibitem[{Moen et~al.(2019)Moen, Bannon, Kudo, Graf, Covert and
  Van~Valen}]{moen2019deep}
\bibinfo{author}{Moen, E.}, \bibinfo{author}{Bannon, D.},
  \bibinfo{author}{Kudo, T.}, \bibinfo{author}{Graf, W.},
  \bibinfo{author}{Covert, M.}, \bibinfo{author}{Van~Valen, D.},
  \bibinfo{year}{2019}.
\newblock \bibinfo{title}{Deep learning for cellular image analysis}.
\newblock \bibinfo{journal}{Nature methods} , \bibinfo{pages}{1--14}.
\bibitem[{Mormont et~al.(2018)Mormont, Geurts and
  Mar{\'e}e}]{mormont2018comparison}
\bibinfo{author}{Mormont, R.}, \bibinfo{author}{Geurts, P.},
  \bibinfo{author}{Mar{\'e}e, R.}, \bibinfo{year}{2018}.
\newblock \bibinfo{title}{Comparison of deep transfer learning strategies for
  digital pathology}, in: \bibinfo{booktitle}{Proceedings of the IEEE
  Conference on Computer Vision and Pattern Recognition Workshops}, pp.
  \bibinfo{pages}{2262--2271}.
\bibitem[{Mundt et~al.(2020)Mundt, Hong, Pliushch and
  Ramesh}]{mundt2020wholistic}
\bibinfo{author}{Mundt, M.}, \bibinfo{author}{Hong, Y.W.},
  \bibinfo{author}{Pliushch, I.}, \bibinfo{author}{Ramesh, V.},
  \bibinfo{year}{2020}.
\newblock \bibinfo{title}{A wholistic view of continual learning with deep
  neural networks: Forgotten lessons and the bridge to active and open world
  learning}.
\newblock \bibinfo{journal}{arXiv preprint arXiv:2009.01797} .
\bibitem[{Munjal et~al.(2020)Munjal, Hayat, Hayat, Sourati and
  Khan}]{munjal2020towards}
\bibinfo{author}{Munjal, P.}, \bibinfo{author}{Hayat, N.},
  \bibinfo{author}{Hayat, M.}, \bibinfo{author}{Sourati, J.},
  \bibinfo{author}{Khan, S.}, \bibinfo{year}{2020}.
\newblock \bibinfo{title}{Towards robust and reproducible active learning using
  neural networks}.
\newblock \bibinfo{journal}{ArXiv, abs/2002.09564} .
\bibitem[{Ozdemir et~al.(2018)Ozdemir, Peng, Tanner, Fuernstahl and
  Goksel}]{ozdemir2018active}
\bibinfo{author}{Ozdemir, F.}, \bibinfo{author}{Peng, Z.},
  \bibinfo{author}{Tanner, C.}, \bibinfo{author}{Fuernstahl, P.},
  \bibinfo{author}{Goksel, O.}, \bibinfo{year}{2018}.
\newblock \bibinfo{title}{Active learning for segmentation by optimizing
  content information for maximal entropy}, in: \bibinfo{booktitle}{Deep
  Learning in Medical Image Analysis and Multimodal Learning for Clinical
  Decision Support}. \bibinfo{publisher}{Springer}, pp.
  \bibinfo{pages}{183--191}.
\bibitem[{Parisi et~al.(2019)Parisi, Kemker, Part, Kanan and
  Wermter}]{parisi2019continual}
\bibinfo{author}{Parisi, G.I.}, \bibinfo{author}{Kemker, R.},
  \bibinfo{author}{Part, J.L.}, \bibinfo{author}{Kanan, C.},
  \bibinfo{author}{Wermter, S.}, \bibinfo{year}{2019}.
\newblock \bibinfo{title}{Continual lifelong learning with neural networks: A
  review}.
\newblock \bibinfo{journal}{Neural Networks} \bibinfo{volume}{113},
  \bibinfo{pages}{54--71}.
\bibitem[{Pathak et~al.(2016)Pathak, Krahenbuhl, Donahue, Darrell and
  Efros}]{pathak2016context}
\bibinfo{author}{Pathak, D.}, \bibinfo{author}{Krahenbuhl, P.},
  \bibinfo{author}{Donahue, J.}, \bibinfo{author}{Darrell, T.},
  \bibinfo{author}{Efros, A.A.}, \bibinfo{year}{2016}.
\newblock \bibinfo{title}{Context encoders: Feature learning by inpainting},
  in: \bibinfo{booktitle}{Proceedings of the IEEE Conference on Computer Vision
  and Pattern Recognition}, pp. \bibinfo{pages}{2536--2544}.
\bibitem[{Purushwalkam and Gupta(2020)}]{purushwalkam2020demystifying}
\bibinfo{author}{Purushwalkam, S.}, \bibinfo{author}{Gupta, A.},
  \bibinfo{year}{2020}.
\newblock \bibinfo{title}{Demystifying contrastive self-supervised learning:
  Invariances, augmentations and dataset biases}.
\newblock \bibinfo{journal}{arXiv preprint arXiv:2007.13916} .
\bibitem[{Ravizza et~al.(2019)Ravizza, Huschto, Adamov, B{\"o}hm, B{\"u}sser,
  Fl{\"o}ther, Hinzmann, K{\"o}nig, McAhren, Robertson
  et~al.}]{ravizza2019predicting}
\bibinfo{author}{Ravizza, S.}, \bibinfo{author}{Huschto, T.},
  \bibinfo{author}{Adamov, A.}, \bibinfo{author}{B{\"o}hm, L.},
  \bibinfo{author}{B{\"u}sser, A.}, \bibinfo{author}{Fl{\"o}ther, F.F.},
  \bibinfo{author}{Hinzmann, R.}, \bibinfo{author}{K{\"o}nig, H.},
  \bibinfo{author}{McAhren, S.M.}, \bibinfo{author}{Robertson, D.H.}, et~al.,
  \bibinfo{year}{2019}.
\newblock \bibinfo{title}{Predicting the early risk of chronic kidney disease
  in patients with diabetes using real-world data}.
\newblock \bibinfo{journal}{Nature medicine} \bibinfo{volume}{25},
  \bibinfo{pages}{57--59}.
\bibitem[{Ren et~al.(2020)Ren, Xiao, Chang, Huang, Li, Chen and
  Wang}]{ren2020survey}
\bibinfo{author}{Ren, P.}, \bibinfo{author}{Xiao, Y.}, \bibinfo{author}{Chang,
  X.}, \bibinfo{author}{Huang, P.Y.}, \bibinfo{author}{Li, Z.},
  \bibinfo{author}{Chen, X.}, \bibinfo{author}{Wang, X.}, \bibinfo{year}{2020}.
\newblock \bibinfo{title}{A survey of deep active learning}.
\newblock \bibinfo{journal}{arXiv preprint arXiv:2009.00236} .
\bibitem[{Sabour et~al.(2017)Sabour, Frosst and Hinton}]{sabour2017dynamic}
\bibinfo{author}{Sabour, S.}, \bibinfo{author}{Frosst, N.},
  \bibinfo{author}{Hinton, G.E.}, \bibinfo{year}{2017}.
\newblock \bibinfo{title}{Dynamic routing between capsules}.
\newblock \bibinfo{journal}{arXiv preprint arXiv:1710.09829} .
\bibitem[{Scheffer et~al.(2001)Scheffer, Decomain and
  Wrobel}]{scheffer2001active}
\bibinfo{author}{Scheffer, T.}, \bibinfo{author}{Decomain, C.},
  \bibinfo{author}{Wrobel, S.}, \bibinfo{year}{2001}.
\newblock \bibinfo{title}{Active hidden markov models for information
  extraction}, in: \bibinfo{booktitle}{International Symposium on Intelligent
  Data Analysis}, \bibinfo{organization}{Springer}. pp.
  \bibinfo{pages}{309--318}.
\bibitem[{Sener and Savarese(2017)}]{sener2017active}
\bibinfo{author}{Sener, O.}, \bibinfo{author}{Savarese, S.},
  \bibinfo{year}{2017}.
\newblock \bibinfo{title}{Active learning for convolutional neural networks: A
  core-set approach}.
\newblock \bibinfo{journal}{arXiv preprint arXiv:1708.00489} .
\bibitem[{Settles()}]{settles2010active}
\bibinfo{author}{Settles, B.}, .
\newblock \bibinfo{title}{Active learning literature survey}.
\newblock \bibinfo{journal}{University of Wisconsin, Madison}
  \bibinfo{volume}{52}, \bibinfo{pages}{11}.
\bibitem[{Shannon(1948)}]{shannon1948mathematical}
\bibinfo{author}{Shannon, C.E.}, \bibinfo{year}{1948}.
\newblock \bibinfo{title}{A mathematical theory of communication}.
\newblock \bibinfo{journal}{Bell system technical journal}
  \bibinfo{volume}{27}, \bibinfo{pages}{379--423}.
\bibitem[{Shao et~al.(2018)Shao, Sun and Zhang}]{shao2018deep}
\bibinfo{author}{Shao, W.}, \bibinfo{author}{Sun, L.}, \bibinfo{author}{Zhang,
  D.}, \bibinfo{year}{2018}.
\newblock \bibinfo{title}{Deep active learning for nucleus classification in
  pathology images}, in: \bibinfo{booktitle}{2018 IEEE 15th International
  Symposium on Biomedical Imaging (ISBI 2018)}, \bibinfo{organization}{IEEE}.
  pp. \bibinfo{pages}{199--202}.
\bibitem[{Shen et~al.(2019)Shen, Liu, Peters, Staib, Essert, Zhou, Yap and
  Khan}]{shen2019medical}
\bibinfo{author}{Shen, D.}, \bibinfo{author}{Liu, T.}, \bibinfo{author}{Peters,
  T.M.}, \bibinfo{author}{Staib, L.H.}, \bibinfo{author}{Essert, C.},
  \bibinfo{author}{Zhou, S.}, \bibinfo{author}{Yap, P.T.},
  \bibinfo{author}{Khan, A.}, \bibinfo{year}{2019}.
\newblock \bibinfo{title}{Medical Image Computing and Computer Assisted
  Intervention--MICCAI 2019: 22nd International Conference, Shenzhen, China,
  October 13--17, 2019, Proceedings}. volume \bibinfo{volume}{11767}.
\newblock \bibinfo{publisher}{Springer Nature}.
\bibitem[{Shui et~al.(2020)Shui, Zhou, Gagn{\'e} and Wang}]{shui2020deep}
\bibinfo{author}{Shui, C.}, \bibinfo{author}{Zhou, F.},
  \bibinfo{author}{Gagn{\'e}, C.}, \bibinfo{author}{Wang, B.},
  \bibinfo{year}{2020}.
\newblock \bibinfo{title}{Deep active learning: Unified and principled method
  for query and training}, in: \bibinfo{booktitle}{International Conference on
  Artificial Intelligence and Statistics}, \bibinfo{organization}{PMLR}. pp.
  \bibinfo{pages}{1308--1318}.
\bibitem[{Simonyan and Zisserman(2014)}]{simonyan2014very}
\bibinfo{author}{Simonyan, K.}, \bibinfo{author}{Zisserman, A.},
  \bibinfo{year}{2014}.
\newblock \bibinfo{title}{Very deep convolutional networks for large-scale
  image recognition}.
\newblock \bibinfo{journal}{arXiv preprint arXiv:1409.1556} .
\bibitem[{Sourati et~al.(2016)Sourati, Akcakaya, Dy, Leen and
  Erdogmus}]{sourati2016classification}
\bibinfo{author}{Sourati, J.}, \bibinfo{author}{Akcakaya, M.},
  \bibinfo{author}{Dy, J.G.}, \bibinfo{author}{Leen, T.K.},
  \bibinfo{author}{Erdogmus, D.}, \bibinfo{year}{2016}.
\newblock \bibinfo{title}{Classification active learning based on mutual
  information}.
\newblock \bibinfo{journal}{Entropy} \bibinfo{volume}{18}, \bibinfo{pages}{51}.
\bibitem[{Sourati et~al.(2018)Sourati, Gholipour, Dy, Kurugol and
  Warfield}]{sourati2018active}
\bibinfo{author}{Sourati, J.}, \bibinfo{author}{Gholipour, A.},
  \bibinfo{author}{Dy, J.G.}, \bibinfo{author}{Kurugol, S.},
  \bibinfo{author}{Warfield, S.K.}, \bibinfo{year}{2018}.
\newblock \bibinfo{title}{Active deep learning with fisher information for
  patch-wise semantic segmentation}, in: \bibinfo{booktitle}{Deep Learning in
  Medical Image Analysis and Multimodal Learning for Clinical Decision
  Support}. \bibinfo{publisher}{Springer}, pp. \bibinfo{pages}{83--91}.
\bibitem[{Sourati et~al.(2019)Sourati, Gholipour, Dy, Tomas-Fernandez, Kurugol
  and Warfield}]{sourati2019intelligent}
\bibinfo{author}{Sourati, J.}, \bibinfo{author}{Gholipour, A.},
  \bibinfo{author}{Dy, J.G.}, \bibinfo{author}{Tomas-Fernandez, X.},
  \bibinfo{author}{Kurugol, S.}, \bibinfo{author}{Warfield, S.K.},
  \bibinfo{year}{2019}.
\newblock \bibinfo{title}{Intelligent labeling based on fisher information for
  medical image segmentation using deep learning}.
\newblock \bibinfo{journal}{IEEE transactions on medical imaging}
  \bibinfo{volume}{38}, \bibinfo{pages}{2642--2653}.
\bibitem[{Szegedy et~al.(2015)Szegedy, Liu, Jia, Sermanet, Reed, Anguelov,
  Erhan, Vanhoucke, Rabinovich et~al.}]{szegedy2015going}
\bibinfo{author}{Szegedy, C.}, \bibinfo{author}{Liu, W.}, \bibinfo{author}{Jia,
  Y.}, \bibinfo{author}{Sermanet, P.}, \bibinfo{author}{Reed, S.},
  \bibinfo{author}{Anguelov, D.}, \bibinfo{author}{Erhan, D.},
  \bibinfo{author}{Vanhoucke, V.}, \bibinfo{author}{Rabinovich, A.}, et~al.,
  \bibinfo{year}{2015}.
\newblock \bibinfo{title}{Going deeper with convolutions},
  \bibinfo{organization}{Proceedings of the IEEE Conference on Computer Vision
  and Pattern Recognition}.
\bibitem[{Tajbakhsh et~al.(2015)Tajbakhsh, Gotway and
  Liang}]{tajbakhsh2015computer}
\bibinfo{author}{Tajbakhsh, N.}, \bibinfo{author}{Gotway, M.B.},
  \bibinfo{author}{Liang, J.}, \bibinfo{year}{2015}.
\newblock \bibinfo{title}{Computer-aided pulmonary embolism detection using a
  novel vessel-aligned multi-planar image representation and convolutional
  neural networks}, in: \bibinfo{booktitle}{International Conference on Medical
  Image Computing and Computer-Assisted Intervention},
  \bibinfo{organization}{Springer}. pp. \bibinfo{pages}{62--69}.
\bibitem[{Tajbakhsh et~al.(2020)Tajbakhsh, Jeyaseelan, Li, Chiang, Wu and
  Ding}]{tajbakhsh2020embracing}
\bibinfo{author}{Tajbakhsh, N.}, \bibinfo{author}{Jeyaseelan, L.},
  \bibinfo{author}{Li, Q.}, \bibinfo{author}{Chiang, J.N.},
  \bibinfo{author}{Wu, Z.}, \bibinfo{author}{Ding, X.}, \bibinfo{year}{2020}.
\newblock \bibinfo{title}{Embracing imperfect datasets: A review of deep
  learning solutions for medical image segmentation}.
\newblock \bibinfo{journal}{Medical Image Analysis} , \bibinfo{pages}{101693}.
\bibitem[{Tajbakhsh et~al.(2019)Tajbakhsh, Shin, Gotway and
  Liang}]{tajbakhsh2019computer}
\bibinfo{author}{Tajbakhsh, N.}, \bibinfo{author}{Shin, J.Y.},
  \bibinfo{author}{Gotway, M.B.}, \bibinfo{author}{Liang, J.},
  \bibinfo{year}{2019}.
\newblock \bibinfo{title}{Computer-aided detection and visualization of
  pulmonary embolism using a novel, compact, and discriminative image
  representation}.
\newblock \bibinfo{journal}{Medical image analysis} \bibinfo{volume}{58},
  \bibinfo{pages}{101541}.
\bibitem[{Tajbakhsh et~al.(2016)Tajbakhsh, Shin, Gurudu, Hurst, Kendall, Gotway
  and Liang}]{tajbakhsh2016convolutional}
\bibinfo{author}{Tajbakhsh, N.}, \bibinfo{author}{Shin, J.Y.},
  \bibinfo{author}{Gurudu, S.R.}, \bibinfo{author}{Hurst, R.T.},
  \bibinfo{author}{Kendall, C.B.}, \bibinfo{author}{Gotway, M.B.},
  \bibinfo{author}{Liang, J.}, \bibinfo{year}{2016}.
\newblock \bibinfo{title}{Convolutional neural networks for medical image
  analysis: Full training or fine tuning?}
\newblock \bibinfo{journal}{IEEE transactions on medical imaging}
  \bibinfo{volume}{35}, \bibinfo{pages}{1299--1312}.
\bibitem[{Tang et~al.(2018)Tang, Wang, Harrison, Lu, Xiao and
  Summers}]{tang2018attention}
\bibinfo{author}{Tang, Y.}, \bibinfo{author}{Wang, X.},
  \bibinfo{author}{Harrison, A.P.}, \bibinfo{author}{Lu, L.},
  \bibinfo{author}{Xiao, J.}, \bibinfo{author}{Summers, R.M.},
  \bibinfo{year}{2018}.
\newblock \bibinfo{title}{Attention-guided curriculum learning for weakly
  supervised classification and localization of thoracic diseases on chest
  radiographs}, in: \bibinfo{booktitle}{International Workshop on Machine
  Learning in Medical Imaging}, \bibinfo{organization}{Springer}. pp.
  \bibinfo{pages}{249--258}.
\bibitem[{Touvron et~al.(2020)Touvron, Vedaldi, Douze and
  J{\'e}gou}]{touvron2020fixing}
\bibinfo{author}{Touvron, H.}, \bibinfo{author}{Vedaldi, A.},
  \bibinfo{author}{Douze, M.}, \bibinfo{author}{J{\'e}gou, H.},
  \bibinfo{year}{2020}.
\newblock \bibinfo{title}{Fixing the train-test resolution discrepancy:
  Fixefficientnet}.
\newblock \bibinfo{journal}{arXiv preprint arXiv:2003.08237} .
\bibitem[{Tsymbalov et~al.(2018)Tsymbalov, Panov and
  Shapeev}]{tsymbalov2018dropout}
\bibinfo{author}{Tsymbalov, E.}, \bibinfo{author}{Panov, M.},
  \bibinfo{author}{Shapeev, A.}, \bibinfo{year}{2018}.
\newblock \bibinfo{title}{Dropout-based active learning for regression}, in:
  \bibinfo{booktitle}{International conference on analysis of images, social
  networks and texts}, \bibinfo{organization}{Springer}. pp.
  \bibinfo{pages}{247--258}.
\bibitem[{Venturini et~al.(2020)Venturini, Papageorghiou, Noble and
  Namburete}]{venturini2020uncertainty}
\bibinfo{author}{Venturini, L.}, \bibinfo{author}{Papageorghiou, A.T.},
  \bibinfo{author}{Noble, J.A.}, \bibinfo{author}{Namburete, A.I.},
  \bibinfo{year}{2020}.
\newblock \bibinfo{title}{Uncertainty estimates as data selection criteria to
  boost omni-supervised learning}, in: \bibinfo{booktitle}{International
  Conference on Medical Image Computing and Computer-Assisted Intervention},
  \bibinfo{organization}{Springer}. pp. \bibinfo{pages}{689--698}.
\bibitem[{Wang et~al.(2018)Wang, Lu, Wu, Chen, Chen and Wu}]{wang2018deep}
\bibinfo{author}{Wang, W.}, \bibinfo{author}{Lu, Y.}, \bibinfo{author}{Wu, B.},
  \bibinfo{author}{Chen, T.}, \bibinfo{author}{Chen, D.Z.},
  \bibinfo{author}{Wu, J.}, \bibinfo{year}{2018}.
\newblock \bibinfo{title}{Deep active self-paced learning for accurate
  pulmonary nodule segmentation}, in: \bibinfo{booktitle}{International
  Conference on Medical Image Computing and Computer-Assisted Intervention},
  \bibinfo{organization}{Springer}. pp. \bibinfo{pages}{723--731}.
\bibitem[{Yamamoto et~al.(2019)Yamamoto, Tsuzuki, Akatsuka, Ueki, Morikawa,
  Numata, Takahara, Tsuyuki, Tsutsumi, Nakazawa et~al.}]{yamamoto2019automated}
\bibinfo{author}{Yamamoto, Y.}, \bibinfo{author}{Tsuzuki, T.},
  \bibinfo{author}{Akatsuka, J.}, \bibinfo{author}{Ueki, M.},
  \bibinfo{author}{Morikawa, H.}, \bibinfo{author}{Numata, Y.},
  \bibinfo{author}{Takahara, T.}, \bibinfo{author}{Tsuyuki, T.},
  \bibinfo{author}{Tsutsumi, K.}, \bibinfo{author}{Nakazawa, R.}, et~al.,
  \bibinfo{year}{2019}.
\newblock \bibinfo{title}{Automated acquisition of explainable knowledge from
  unannotated histopathology images}.
\newblock \bibinfo{journal}{Nature communications} \bibinfo{volume}{10},
  \bibinfo{pages}{1--9}.
\bibitem[{Yang et~al.(2017)Yang, Zhang, Chen, Zhang and
  Chen}]{yang2017suggestive}
\bibinfo{author}{Yang, L.}, \bibinfo{author}{Zhang, Y.}, \bibinfo{author}{Chen,
  J.}, \bibinfo{author}{Zhang, S.}, \bibinfo{author}{Chen, D.Z.},
  \bibinfo{year}{2017}.
\newblock \bibinfo{title}{Suggestive annotation: A deep active learning
  framework for biomedical image segmentation}.
\newblock \bibinfo{journal}{arXiv preprint arXiv:1706.04737} .
\bibitem[{Yu et~al.(2019)Yu, Wang, Li, Fu and Heng}]{yu2019uncertainty}
\bibinfo{author}{Yu, L.}, \bibinfo{author}{Wang, S.}, \bibinfo{author}{Li, X.},
  \bibinfo{author}{Fu, C.W.}, \bibinfo{author}{Heng, P.A.},
  \bibinfo{year}{2019}.
\newblock \bibinfo{title}{Uncertainty-aware self-ensembling model for
  semi-supervised 3d left atrium segmentation}, in:
  \bibinfo{booktitle}{International Conference on Medical Image Computing and
  Computer-Assisted Intervention}, \bibinfo{organization}{Springer}. pp.
  \bibinfo{pages}{605--613}.
\bibitem[{Yuan et~al.(2020)Yuan, Lin and Boyd-Graber}]{yuan2020cold}
\bibinfo{author}{Yuan, M.}, \bibinfo{author}{Lin, H.T.},
  \bibinfo{author}{Boyd-Graber, J.}, \bibinfo{year}{2020}.
\newblock \bibinfo{title}{Cold-start active learning through self-supervised
  language modeling}.
\newblock \bibinfo{journal}{arXiv preprint arXiv:2010.09535} .
\bibitem[{Yuan and Zhang(2013)}]{yuan2013truncated}
\bibinfo{author}{Yuan, X.T.}, \bibinfo{author}{Zhang, T.},
  \bibinfo{year}{2013}.
\newblock \bibinfo{title}{Truncated power method for sparse eigenvalue
  problems}.
\newblock \bibinfo{journal}{Journal of Machine Learning Research}
  \bibinfo{volume}{14}, \bibinfo{pages}{899--925}.
\bibitem[{Zhang et~al.(2016)Zhang, Isola and Efros}]{zhang2016colorful}
\bibinfo{author}{Zhang, R.}, \bibinfo{author}{Isola, P.},
  \bibinfo{author}{Efros, A.A.}, \bibinfo{year}{2016}.
\newblock \bibinfo{title}{Colorful image colorization}, in:
  \bibinfo{booktitle}{Proceedings of the European Conference on Computer
  Vision}, \bibinfo{organization}{Springer}. pp. \bibinfo{pages}{649--666}.
\bibitem[{Zhou et~al.(2017a)Zhou, Lapedriza, Khosla, Oliva and
  Torralba}]{zhou2017places}
\bibinfo{author}{Zhou, B.}, \bibinfo{author}{Lapedriza, A.},
  \bibinfo{author}{Khosla, A.}, \bibinfo{author}{Oliva, A.},
  \bibinfo{author}{Torralba, A.}, \bibinfo{year}{2017}a.
\newblock \bibinfo{title}{Places: A 10 million image database for scene
  recognition}.
\newblock \bibinfo{journal}{IEEE transactions on pattern analysis and machine
  intelligence} .
\bibitem[{Zhou et~al.(2019a)Zhou, Rueckert and Fichtinger}]{zhou2019handbook}
\bibinfo{author}{Zhou, S.K.}, \bibinfo{author}{Rueckert, D.},
  \bibinfo{author}{Fichtinger, G.}, \bibinfo{year}{2019}a.
\newblock \bibinfo{title}{Handbook of medical image computing and computer
  assisted intervention}.
\newblock \bibinfo{publisher}{Academic Press}.
\bibitem[{Zhou et~al.(2019b)Zhou, Shin, Feng, Hurst, Kendall and
  Liang}]{zhou2019integrating}
\bibinfo{author}{Zhou, Z.}, \bibinfo{author}{Shin, J.}, \bibinfo{author}{Feng,
  R.}, \bibinfo{author}{Hurst, R.T.}, \bibinfo{author}{Kendall, C.B.},
  \bibinfo{author}{Liang, J.}, \bibinfo{year}{2019}b.
\newblock \bibinfo{title}{Integrating active learning and transfer learning for
  carotid intima-media thickness video interpretation}.
\newblock \bibinfo{journal}{Journal of digital imaging} \bibinfo{volume}{32},
  \bibinfo{pages}{290--299}.
\bibitem[{Zhou et~al.(2017b)Zhou, Shin, Zhang, Gurudu, Gotway and
  Liang}]{zhou2017fine}
\bibinfo{author}{Zhou, Z.}, \bibinfo{author}{Shin, J.}, \bibinfo{author}{Zhang,
  L.}, \bibinfo{author}{Gurudu, S.}, \bibinfo{author}{Gotway, M.},
  \bibinfo{author}{Liang, J.}, \bibinfo{year}{2017}b.
\newblock \bibinfo{title}{Fine-tuning convolutional neural networks for
  biomedical image analysis: actively and incrementally}, in:
  \bibinfo{booktitle}{Proceedings of the IEEE Conference on Computer Vision and
  Pattern Recognition}, pp. \bibinfo{pages}{7340--7349}.
\bibitem[{Zhou et~al.(2021)Zhou, Sodha, Pang, Gotway and
  Liang}]{zhou2021models}
\bibinfo{author}{Zhou, Z.}, \bibinfo{author}{Sodha, V.}, \bibinfo{author}{Pang,
  J.}, \bibinfo{author}{Gotway, M.B.}, \bibinfo{author}{Liang, J.},
  \bibinfo{year}{2021}.
\newblock \bibinfo{title}{Models genesis}.
\newblock \bibinfo{journal}{Medical Image Analysis} \bibinfo{volume}{67},
  \bibinfo{pages}{101840}.
\newblock \URLprefix
  \url{http://www.sciencedirect.com/science/article/pii/S1361841520302048},
  \DOIprefix\doi{https://doi.org/10.1016/j.media.2020.101840}.
\bibitem[{Zhou et~al.(2019c)Zhou, Sodha, Rahman~Siddiquee, Feng, Tajbakhsh,
  Gotway and Liang}]{zhou2019models}
\bibinfo{author}{Zhou, Z.}, \bibinfo{author}{Sodha, V.},
  \bibinfo{author}{Rahman~Siddiquee, M.M.}, \bibinfo{author}{Feng, R.},
  \bibinfo{author}{Tajbakhsh, N.}, \bibinfo{author}{Gotway, M.B.},
  \bibinfo{author}{Liang, J.}, \bibinfo{year}{2019}c.
\newblock \bibinfo{title}{Models genesis: Generic autodidactic models for 3d
  medical image analysis}, in: \bibinfo{booktitle}{Medical Image Computing and
  Computer Assisted Intervention -- MICCAI 2019}, \bibinfo{publisher}{Springer
  International Publishing}, \bibinfo{address}{Cham}. pp.
  \bibinfo{pages}{384--393}.
\newblock \URLprefix
  \url{https://link.springer.com/chapter/10.1007/978-3-030-32251-9_42}.
\bibitem[{Zhu et~al.(2020)Zhu, Li, Hu, Ma, Zhou and Zheng}]{zhu2020rubik}
\bibinfo{author}{Zhu, J.}, \bibinfo{author}{Li, Y.}, \bibinfo{author}{Hu, Y.},
  \bibinfo{author}{Ma, K.}, \bibinfo{author}{Zhou, S.K.},
  \bibinfo{author}{Zheng, Y.}, \bibinfo{year}{2020}.
\newblock \bibinfo{title}{Rubik’s cube+: A self-supervised feature learning
  framework for 3d medical image analysis}.
\newblock \bibinfo{journal}{Medical Image Analysis} \bibinfo{volume}{64},
  \bibinfo{pages}{101746}.

\end{thebibliography}

\newpage
\appendix

\newpage
\section{Selected Images Gallery}
\label{sec:image_gallery}

\begin{figure}[h]
\begin{center}
\includegraphics[width=1.0\linewidth]{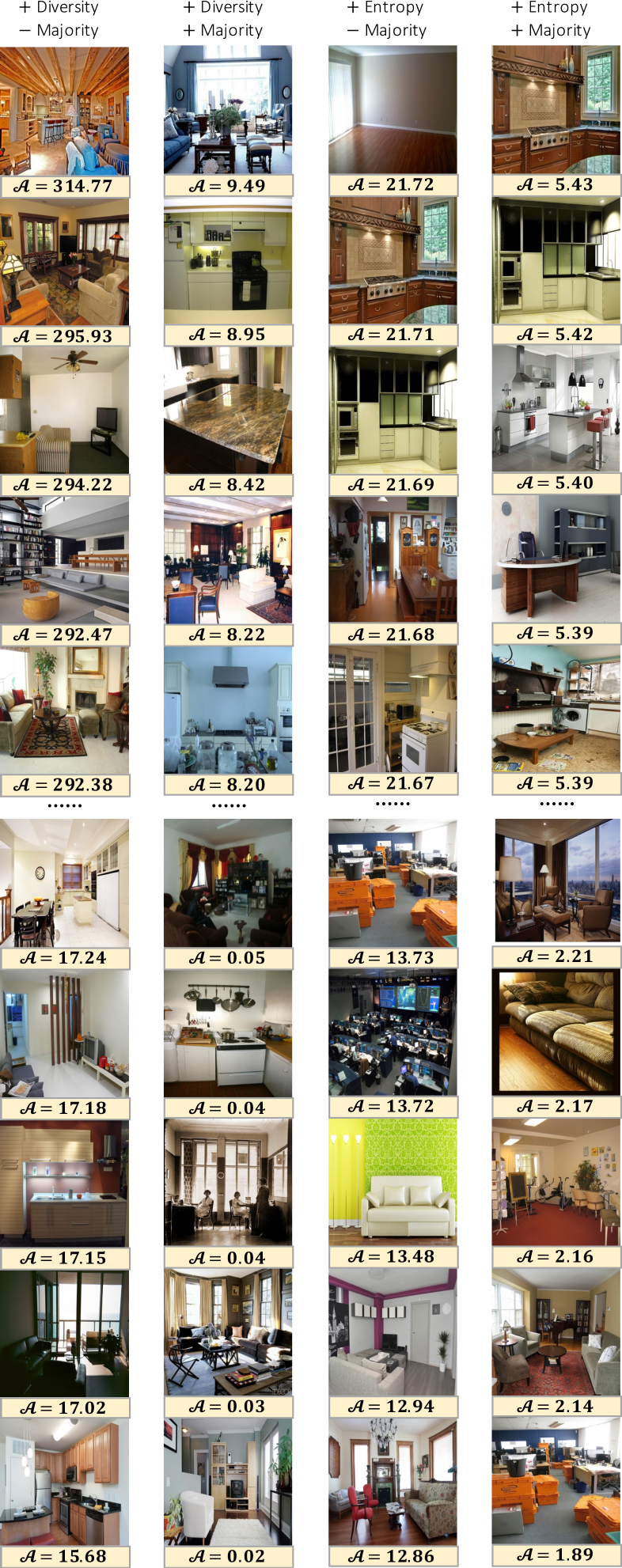}
\end{center}
\caption{Gallery of top five and bottom five candidates actively selected at Step 11 by the methods proposed in Sec.~\ref{sec:entropy_diversity} and Sec~\ref{sec:majority_selection} under the experimental setting.}
\label{fig:selected_image_gallary}
\end{figure}

We illustrate the top and bottom five images selected by four active selection strategies (\ie diversity, diversity+majority, entropy and entropy+majority) from \textsc{Places-3} at Step 11 in \figurename~\ref{fig:selected_image_gallary} to create a visual impression of the appearance of newly selected images. Such a gallery offers an intuitive way to analyze the most/least favored images and has helped us develop different active selection strategies.

\begin{figure}[h]
\begin{center}
\includegraphics[width=1.0\linewidth]{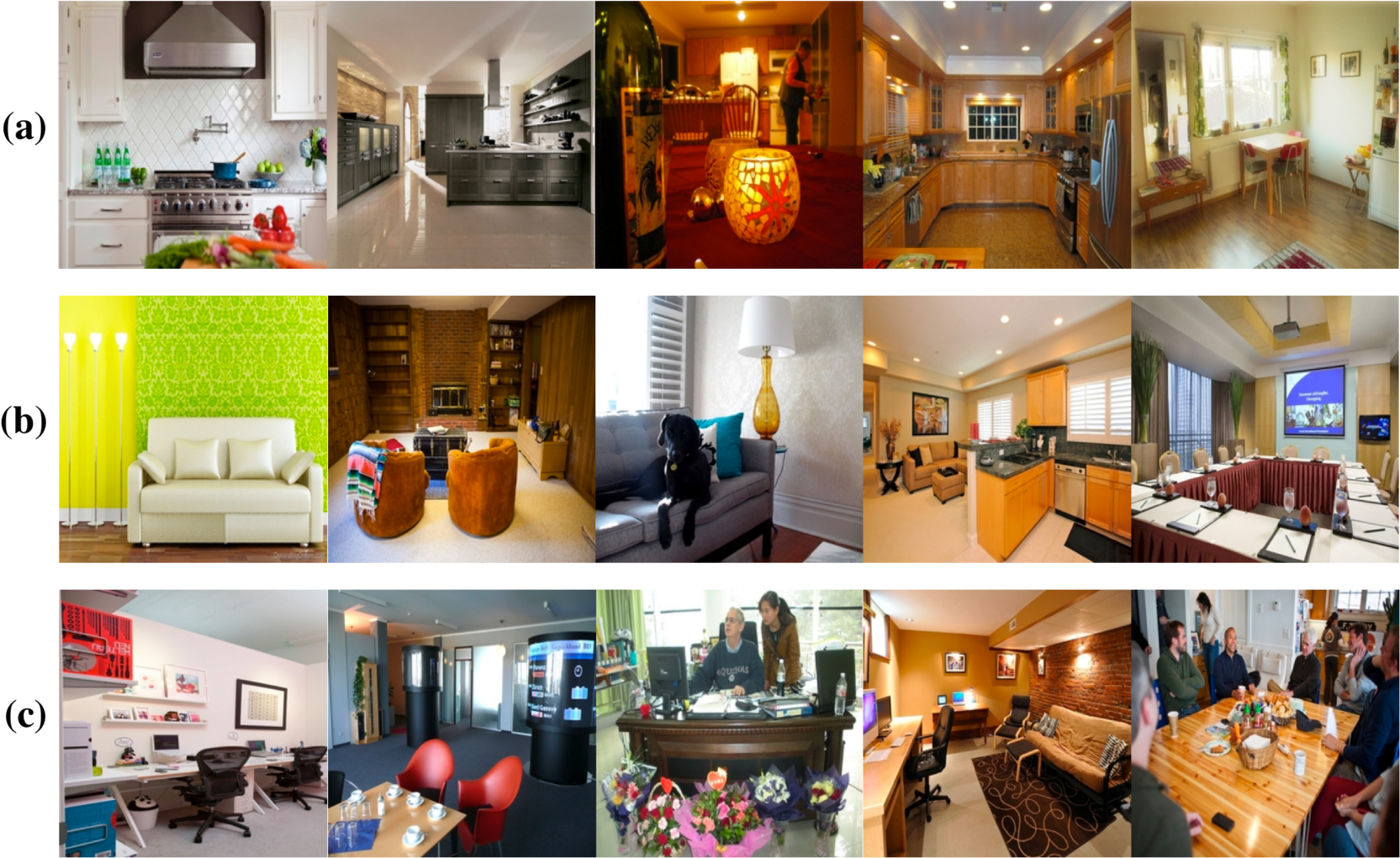}
\end{center}
\caption{We illustrate the ideas behind ACFT by utilizing \textsc{Places-3}~\citep{zhou2017places} for scene classification in natural images. For simplicity yet without loss of generality, we limit to 3 categories: (a)~{\em kitchen}, (b)~{\em living room}, and (c)~{\em office}. \textsc{Places-3} has 15,100 images in each category.}
\label{fig:places_database}
\end{figure}

\end{document}